\documentclass[submission,copyright,creativecommons,sharealike]{eptcs}
\providecommand{\event}{ACT 2020} %
\usepackage{breakurl}             %
\usepackage{underscore}           %

\title{Cyber Kittens, or \\ Some First Steps Towards Categorical Cybernetics}
\author{Toby St. Clere Smithe
  \institute{Department of Experimental Psychology, \\
    University of Oxford}
  \email{arxiv@tsmithe.net}
}
\date{\today}

\usepackage[utf8]{inputenc}
\usepackage[T1]{fontenc}
\usepackage{graphicx}
\usepackage{grffile}
\usepackage{longtable}
\usepackage{wrapfig}
\usepackage{rotating}
\usepackage[normalem]{ulem}
\usepackage{amsmath}
\usepackage{textcomp}
\usepackage{amssymb}
\usepackage{capt-of}
\usepackage{hyperref}

\pdfoutput=1

\usepackage[utf8]{inputenc}    
\usepackage[T1]{fontenc}

\usepackage{amsfonts}
\usepackage{amssymb}
\usepackage{amsthm}
\usepackage{amsmath}
\usepackage{caption}
\usepackage[inline]{enumitem}
\setlist[enumerate]{label=(\alph*)}
\usepackage{etoolbox}
\usepackage{stmaryrd}
\usepackage[dvipsnames]{xcolor}
\usepackage{ifthen}
\usepackage{suffix}
\usepackage{verbatim} %

\usepackage{xfrac}

\usepackage{bm}
\usepackage{physics}
\usepackage[safe]{tipa} %
\usepackage{mathtools} %
\usepackage{color}
\usepackage{wrapfig}
\usepackage{tikz}
\usepackage{tikzit}

\usepackage[framemethod=tikz]{mdframed}

\newcommand*{\secref}[1]{\S\ref{#1}}

\usepackage[numbers,square]{natbib}

\usepackage{hyperref}
\hypersetup{colorlinks,linkcolor=darkblue,citecolor=darkblue,urlcolor=darkblue,breaklinks=true,pdfusetitle}

\usepackage[capitalize]{cleveref} %

\usepackage{graphicx}

\usepackage{pict2e}

\def\mdash{{\hbox{-}}}

\makeatletter
\newcommand{\adjunction}{\@ifstar\named@adjunction\normal@adjunction}
\newcommand{\normal@adjunction}[4]{%
  #1\colon #2%
  \mathrel{\vcenter{%
    \offinterlineskip\m@th
    \ialign{%
      \hfil$##$\hfil\cr
      \longrightharpoonup\cr
      \noalign{\kern-.3ex}
      \smallbot\cr
      \longleftharpoondown\cr
    }%
  }}%
  #3 \noloc #4%
}
\newcommand{\named@adjunction}[4]{%
  #2%
  \mathrel{\vcenter{%
    \offinterlineskip\m@th
    \ialign{%
      \hfil$##$\hfil\cr
      \scriptstyle#1\cr
      \noalign{\kern.1ex}
      \longrightharpoonup\cr
      \noalign{\kern-.3ex}
      \smallbot\cr
      \longleftharpoondown\cr
      \scriptstyle#4\cr
    }%
  }}%
  #3%
}
\newcommand{\longrightharpoonup}{\relbar\joinrel\rightharpoonup}
\newcommand{\longleftharpoondown}{\leftharpoondown\joinrel\relbar}
\newcommand\noloc{%
  \nobreak
  \mspace{6mu plus 1mu}
  {:}
  \nonscript\mkern-\thinmuskip
  \mathpunct{}
  \mspace{2mu}
}
\newcommand{\smallbot}{%
  \begingroup\setlength\unitlength{.15em}%
  \begin{picture}(1,1)
  \roundcap
  \polyline(0,0)(1,0)
  \polyline(0.5,0)(0.5,1)
  \end{picture}%
  \endgroup
}
\makeatother

\newcommand\optar[2]{\ensuremath{\braket{#1\:}{\: #2}}}
\newcommand\bdag[1]{\ensuremath{#1_{(\cdot)}^\dag}}

\newcommand\interp[1]{\ensuremath{\llbracket #1 \rrbracket}}

\newcommand*\circled[3]
  {\tikz[baseline={([yshift=#2]current bounding box.center)}]{
      \node[shape=circle,draw,inner sep=#1] (char) {#3};}}

\newcommand{\docircL}[3]{\mathbin{\circled{#1}{#2}{\scalebox{#3}{\tiny{$\mathsf{L}$}}}}}
\newcommand{\docircR}[3]{\mathbin{\circled{#1}{#2}{\scalebox{#3}{\tiny{$\mathsf{R}$}}}}}

\newcommand*{\circL}{
  \mathchoice{\docircL{0.75pt}{-0.60ex}{1}}
             {\docircL{0.75pt}{-0.60ex}{1}}
             {\docircL{0.3pt}{-0.45ex}{0.70}}
             {\docircL{0.1pt}{-0.35ex}{0.50}}}
\newcommand*{\circR}{
  \mathchoice{\docircR{0.75pt}{-0.60ex}{1}}
             {\docircR{0.75pt}{-0.60ex}{1}}
             {\docircR{0.3pt}{-0.45ex}{0.70}}
             {\docircR{0.1pt}{-0.35ex}{0.50}}}

\let\op=\relax

\DeclareMathOperator{\op}{^\text{op}}

\newcommand{\rr}{{\mathbb{R}}}

\newcommand{\Cat}[1]{\mathbf{#1}}
\newcommand{\cat}[1]{\mathcal{#1}}
\newcommand{\Fun}[1]{\mathsf{#1}}

\newcommand{\Kl}{\mathcal{K}\mspace{-2mu}\ell}

\DeclareMathOperator*{\E}{\mathbb{E}}

\renewcommand{\d}{\mathrm{d}}

\newcommand{\Pow}{\mathcal{P}}

\newcommand{\Dst}{\mathcal{D}}

\newcommand{\Giry}{\mathcal{G}}

\DeclareMathOperator{\id}{\mathsf{id}}

\DeclareMathOperator{\Set}{\Cat{Set}}

\newcommand{\xto}[1]{\xrightarrow{#1}}

\makeatletter
\newcommand{\mathoverlap}[2]{\mathpalette\mathoverlap@{{#1}{#2}}}
\newcommand{\mathoverlap@}[2]{\mathoverlap@@{#1}#2}
\newcommand{\mathoverlap@@}[3]{\ooalign{$\m@th#1#2$\crcr\hidewidth$\m@th#1#3$\hidewidth}}
\makeatother

\newcommand{\klcirc}{\bullet} %
\newcommand*{\smallklcirc}{\raisebox{0.18ex}{\scalebox{0.66}{$\klcirc$}}}
\newcommand{\klto}{\mathoverlap{\rightarrow}{\smallklcirc\,}}

\newcommand{\xklto}[1]{\mathoverlap{\xrightarrow{#1}}{\smallklcirc\,}}

\newcommand{\lenscirc}{
  \mathbin{\mathoverlap{\circ}{\raisebox{0.375ex}{\scalebox{1.0}[0.33]{$|$}}}}
}
\newcommand{\lensto}{\mathrel{\ooalign{\hfil$\mapstochar\mkern5mu$\hfil\cr$\to$\cr}}}

\makeatletter
\providecommand*{\xmapstofill@}{%
  \arrowfill@{\mapstochar\relbar}\relbar\rightarrow
}
\providecommand*{\xmapsto}[2][]{%
  \ext@arrow 0395\xmapstofill@{#1}{#2}%
}

\makeatletter
\def\slashedarrowfill@#1#2#3#4#5{%
  $\m@th\thickmuskip0mu\medmuskip\thickmuskip\thinmuskip\thickmuskip
   \relax#5#1\mkern-7mu%
   \cleaders\hbox{$#5\mkern-2mu#2\mkern-2mu$}\hfill
   \mathclap{#3}\mathclap{#2}%
   \cleaders\hbox{$#5\mkern-2mu#2\mkern-2mu$}\hfill
   \mkern-7mu#4$%
}
\def\rightslashedarrowfill@{%
  \slashedarrowfill@\relbar\relbar\mapstochar\rightarrow}
\newcommand\xslashedrightarrow[2][]{%
  \ext@arrow 0055{\rightslashedarrowfill@}{#1}{#2}}
\makeatother

\theoremstyle{definition}
\newtheorem{defn}{Definition}[section]

\newtheorem{ex}[defn]{Example}

\newtheorem{conjecture}[defn]{Conjecture}

\newtheorem{rmk}[defn]{Remark}
\newtheorem{obs}[defn]{Observation}

\newtheorem{prop}[defn]{Proposition}
\newtheorem{lemma}[defn]{Lemma}
\newtheorem{thm}[defn]{Theorem}
\newtheorem{cor}[defn]{Corollary}

\newtheorem*{thm*}{Theorem}
\newtheorem*{cor*}{Corollary}

\definecolor{darkblue}{rgb}{0,0,0.7}

\tikzstyle{xshiftu}=[shift = {(#1, 0)}]
\tikzstyle{yshiftu}=[shift = {(0, #1)}]

\tikzstyle{dot}=[inner sep=0.25mm,minimum width=1mm,minimum height=1mm,draw,shape=circle,text depth=-0.2mm]
\tikzstyle{white dot}=[dot,fill=white, draw=black]
\tikzstyle{action}=[dot,fill=white,scale=0.667,inner sep=0.5mm]
\tikzstyle{copier}=[dot,fill=white,scale=2.0]
\tikzstyle{black copier}=[dot,fill=black,scale=2.0]

\tikzstyle{box}=[fill=white, draw=black, shape=rectangle]
\tikzstyle{medium box}=[fill=white, draw=black, shape=rectangle, minimum width=1.5cm, minimum height=0.66cm]
\tikzstyle{arrow box}=[fill=white, draw, shape=rectangle,minimum height=5mm,yshift=-0.5mm,minimum width=5mm]

\tikzstyle{effect}=[regular polygon, regular polygon sides=3,draw]
\tikzstyle{state0}=[regular polygon, regular polygon sides=3,draw,shape border rotate=0]
\tikzstyle{state90}=[regular polygon, regular polygon sides=3,draw,shape border rotate=90]
\tikzstyle{state180}=[regular polygon, regular polygon sides=3,draw,shape border rotate=180]
\tikzstyle{state270}=[regular polygon, regular polygon sides=3,draw,shape border rotate=270]

\tikzstyle{scalar}=[diamond,draw,inner sep=1pt]

\tikzstyle{discarder}=[my ground,draw,inner sep=0pt,minimum width=4.2pt,minimum height=11.2pt,anchor=input,rotate=90]
\tikzstyle{discarder0}=[my ground,draw,inner sep=0pt,minimum width=4.2pt,minimum height=11.2pt,anchor=input,rotate=0]

\tikzstyle{pointy1}=[->]
\tikzstyle{midpoint1}=[-, {postaction={decorate,decoration={markings, mark=at position .5 with {\arrow{>}}}}}]
\tikzstyle{midpointy1pointy}=[->, {postaction={decorate,decoration={markings, mark=at position .5 with {\arrow{>}}}}}]
\tikzstyle{dashed1}=[-, dashed]
\tikzstyle{dotted1}=[-, dotted]
\tikzstyle{dash-pointy}=[->, dashed]

\input{strings.tikzdefs}

\ifexternalizetikz\tikzexternaldisable\fi
\newsavebox\sbground
\savebox\sbground{%
  \begin{tikzpicture}[baseline=0pt]
    \draw (0,-.1ex) to (0,.85ex)
    node[ground IEC,draw,anchor=input,inner sep=0pt,
    minimum width=3.15pt,minimum height=8.4pt,rotate=90] {};
  \end{tikzpicture}%
}
\newcommand{\ground}{\mathord{\usebox\sbground}}

\newsavebox\sbcopier
\savebox\sbcopier{%
  \begin{tikzpicture}[baseline=0pt]
    \node[copier,scale=0.7] (a) at (0,3.8pt) {};
    \draw (a) -- +(-90:.21);
    \draw (a) -- +(45:.21);
    \draw (a) -- +(135:.21);
  \end{tikzpicture}}
\newcommand{\copier}{\mathord{\usebox\sbcopier}}
\ifexternalizetikz\tikzexternalenable\fi

\newsavebox\bsbcopier
\savebox\bsbcopier{%
  \begin{tikzpicture}[baseline=0pt]
    \node[black copier,scale=0.7] (a) at (0,3.8pt) {};
    \draw (a) -- +(-90:.21);
    \draw (a) -- +(45:.21);
    \draw (a) -- +(135:.21);
  \end{tikzpicture}}

\ifexternalizetikz\tikzexternalenable\fi

\def\titlerunning{Cyber Kittens}
\def\authorrunning{T. St. Clere Smithe}

\begin{document}
\maketitle

\begin{abstract}
We define a categorical notion of cybernetic system as a dynamical realisation of a generalized open game, along with a coherence condition. We show that this notion captures a wide class of cybernetic systems in computational neuroscience and statistical machine learning, exposes their compositional structure, and gives an abstract justification for the bidirectional structure empirically observed in cortical circuits. Our construction is built on the observation that Bayesian updates compose optically, a fact which we prove along the way, via a fibred category of state-dependent stochastic channels.
\end{abstract}

\section{Introduction}
\label{sec:orgdeb2123}

Those systems that we might classify as living, adaptive, or somehow intelligent all display a fundamental property: they resist or avoid perturbations that would result in their existence becoming unsustainable. This means that they must somehow be able to sense their current state of affairs (\emph{perception}) and respond appropriately (\emph{action}). In particular, an adaptive system should sense the relevant aspects of its current environmental state, and form expectations about the consequences of that state. In general, the interaction with the environment will be stochastic, and the statistically optimal method of `sensing' and prediction is Bayesian inference.

Typically, however, the system has no direct access to the external state, only to sense data that indirectly have external causes. Moreover, sense data are often very high-dimensional, and predicting their consequences is underdetermined. As a result, it is common to assume that successful organisms are imbued with some kind of \emph{generative model} of the process by which external causes generate their sense data. They can then use this model to infer those actions will bring (their beliefs about) their current state closer to those expectations: a process called \emph{active inference}.

Systems such as these are inherently open, and often their internal models and beliefs are supposed to be structured hierarchically---that is, compositionally. The processes of prediction and action sketched here are naturally bidirectional, and indeed our first contribution in the present work is to show that Bayesian inference is abstractly structured as a \emph{category of optics} \citep{Riley2018Categories,Clarke2020Profunctor}, the emerging canonical formalism for (open) bidirectionally structured compositional systems.

The compositional framework of \emph{open games} \citep{Bolt2019Bayesian,Ghani2016Compositional} builds on categories of optics to describe systems of motivated interacting agents, but it is substantially more general than needed for classical game theory: generalized open games naturally describe any bidirectionally structured open systems that can be associated with a measure of fitness. Consequently, such generalized open games provide a natural home for a compositional theory of interacting cybernetic systems, and using our notion of \emph{Bayesian lens}, we characterize a number of canonical statistical models as \emph{statistical games}.

However, mere open games themselves supply no notion of \emph{dynamics} mediating the interactions. We therefore introduce the concept of \emph{dynamical realisation} of an open game (Definition \ref{def:realisation}), as well as a coherence condition that ensures such a realisation behaves as we would expect from a cybernetic system (Definition \ref{def:cyber-sys}). We use these concepts to show that two prominent frameworks for active inference instantiate such categories of cybernetic systems.

\paragraph{Acknowledgements} We thank the organizers of \emph{Applied Category Theory 2020} for the opportunity to present this work, and the anonymous reviewers for helpful comments and questions. We also thank Bruno Gavranović, Jules Hedges, and Neil Ghani for stimulating and insightful conversations, and credit Jules Hedges for observing the correct form of the Bayesian \(\mathsf{update}\) map in discussions at SYCO 6.

\section{Bayesian Updates Compose Optically}
\label{sec:org69acce1}

We begin by proving that Bayesian updates compose according to the `lens' pattern \citep{Foster2007Combinators} that sits at the heart of categories of open games and other `bidirectional' structures. We first show that Bayesian inversions are `vertical' maps in a fibred category of state-dependent channels. The Grothendieck construction of this structure gives a category of lenses. Open games are commonly defined using the more general `optics' pattern \citep{Bolt2019Bayesian}, and so we also show that, under the Yoneda embedding, our category of lenses is equivalently a category of optics.

Throughout the paper, we work in a general category of stochastic channels; abstractly, this corresponds to a \emph{Markov category} \citep{Fritz2019synthetic} or \emph{copy-delete category} \citep{Cho2017Disintegration}. Familiar examples of such categories include \(\Kl(\Dst)\), the Kleisli category of the finitely-supported distribution monad \(\Dst\), and, for `continuous' probabiliy, \(\Kl(\Giry)\), the Kleisli category of the Giry monad. We will write \(c^\dag_\pi := \bdag{c}(\pi)\) to indicate the \textbf{Bayesian inversion} of the channel \(c\) with respect to a state \(\pi\). Then, given some \(y \in Y\), \(c^\dag_\pi (y)\) is a new `posterior' distribution on X. We will call \(c^\dag_\pi(y)\) the \textbf{Bayesian update} of \(\pi\) along \(c\) given \(y\).

For a substantially expanded version of this section, including proofs and background exposition with precise definitions of Bayesian inversion, see the author's \citep{Smithe2020Bayesian}. We will occasionally here refer to definitions or results in that paper.

\begin{defn}[State-indexed categories] \label{def:stat-cat}
Let \((\cat{C}, \otimes, I)\) be a monoidal category enriched in a Cartesian closed category \(\Cat{V}\). Define the \(\cat{C}\text{-state-indexed}\) category \(\Fun{Stat}: \cat{C}\op \to \Cat{V\mdash Cat}\) as follows. 
\begin{align}
\Fun{Stat} \;\; : \;\; \cat{C}\op \; & \to \; \Cat{V\mdash Cat} \nonumber \\
X & \mapsto \Fun{Stat}(X) := \quad \begin{pmatrix*}[l]
& \Fun{Stat}(X)_0 & := \quad \;\;\; \cat{C}_0 \\
& \Fun{Stat}(X)(A, B) & := \quad \;\;\; \Cat{V}(\cat{C}(I, X), \cat{C}(A, B)) \\
\id_A \: : & \Fun{Stat}(x)(A, A) & := \quad 
\left\{ \begin{aligned}
\id_A : & \; \cat{C}(I, X)     \to     \cat{C}(A, A) \\
        & \quad\;\;\: \rho \quad \mapsto \quad \id_A
\end{aligned} \right. \label{eq:stat} \\
\end{pmatrix*} \\ \nonumber \\
f : \cat{C}(Y, X) & \mapsto \begin{pmatrix*}[c]
\Fun{Stat}(f) \; : & \Fun{Stat}(X) & \to & \Fun{Stat}(Y) \vspace*{0.5em} \\
& \Fun{Stat}(X)_0 & = & \Fun{Stat}(Y)_0 \vspace*{0.5em} \\
& \Cat{V}(\cat{C}(I, X), \cat{C}(A, B)) & \to & \Cat{V}(\cat{C}(I, Y), \cat{C}(A, B)) \vspace*{0.125em} \\
& \alpha & \mapsto & f^\ast \alpha : \big( \, \sigma : \cat{C}(I, Y) \, \big) \mapsto \big( \, \alpha(f \klcirc \sigma) : \cat{C}(A, B) \, \big)
\end{pmatrix*} \nonumber
\end{align}
Composition in each fibre \(\Fun{Stat}(X)\) is given by composition in \(\cat{C}\); that is, by the left and right actions of the profunctor \(\Fun{Stat}(X)(-, =) : \cat{C}\op \times \cat{C} \to \Cat{V}\). Explicitly, given \(\alpha : \Cat{V}(\cat{C}(I, X), \cat{C}(A, B))\) and \(\beta : \Cat{V}(\cat{C}(I, X), \cat{C}(B, C))\), their composite is \(\beta \circ \alpha : \Cat{V}(\cat{C}(I, X), \cat{C}(A, C)) : = \rho \mapsto \beta(\rho) \klcirc \alpha(\rho)\). Since \(\Cat{V}\) is Cartesian, there is a canonical copier \(\copier : x \mapsto (x, x)\) on each object, so we can alternatively write \((\beta \circ \alpha)(\rho) = \big(\beta(-) \klcirc \alpha(-)\big) \circ \copier \circ \rho\). Note that we indicate composition in \(\cat{C}\) by \(\klcirc\) and composition in the fibres \(\Fun{Stat}(X)\) by \(\circ\).
\end{defn}

\begin{ex} \label{ex:stat-meas}
Let \(\Cat{V} = \Cat{Meas}\) be a `convenient' (\emph{i.e.}, Cartesian closed) category of measurable spaces, such as the category of quasi-Borel spaces \citep{Heunen2017Convenient}, let \(\Pow : \Cat{Meas} \to \Cat{Meas}\) be a probability monad defined on this category, and let \(\cat{C} = \Kl(\Pow)\) be the Kleisli category of this monad. Its objects are the objects of \(\Cat{Meas}\), and its hom-spaces \(\Kl(\Pow)(A, B)\) are the spaces \(\Cat{Meas}(A, \Pow B)\) \citep{Fritz2019synthetic}. This \(\cat{C}\) is a monoidal category of stochastic channels, whose monoidal unit \(I\) is the space with a single point. Consequently, states of \(X\) are just measures (distributions) in \(\Pow X\). That is, \(\Kl(\Pow)(I, X) \cong \Cat{Meas}(1, \Pow X)\). Instantiating  \(\Fun{Stat}\) in this setting, we obtain:
\begin{align}
\Fun{Stat} \;\; : \;\; \Kl(\Pow)\op \; & \to \; \Cat{V\mdash Cat} \nonumber \\
X & \mapsto \Fun{Stat}(X) := \quad \begin{pmatrix*}[l]
& \Fun{Stat}(X)_0 & := \quad \;\;\; \Cat{Meas}_0 \\
& \Fun{Stat}(X)(A, B) & := \quad \;\;\; \Cat{Meas}(\Pow X, \Cat{Meas}(A, \Pow B)) \\
\id_A \: : & \Fun{Stat}(X)(A, A) & := \quad
\left\{ \begin{aligned}
\id_A : & \; \Pow X     \to     \Cat{Meas}(A, \Pow A) \\
        & \;\;\; \rho \;\;\, \mapsto \quad \eta_A
\end{aligned} \right. \label{eq:stat-kl-d} \\
\end{pmatrix*} \\
c : \Kl(\Pow)(Y, X) & \mapsto \Fun{Stat}(c) \, := \hfill\nonumber
\end{align}
\begin{equation*}
\begin{pmatrix*}[c]
\Fun{Stat}(c) \; : &\Fun{Stat}(X) & \to & \Fun{Stat}(Y) \vspace*{0.5em} \\
& \Fun{Stat}(X)_0 & = & \Fun{Stat}(Y)_0 \vspace*{0.5em} \\
& \begin{pmatrix*}[l]
    d^\dag : & \Pow X & \to \Kl(\Pow)(A, B) \\
    & \; \pi & \mapsto \quad \quad d^\dag_\pi
  \end{pmatrix*}
  & \mapsto &
  \begin{pmatrix*}
    c^\ast d^\dag : \Pow Y \to \Kl(\Pow)(A, B) \\
    \rho \quad \mapsto \quad d^\dag_{c \klcirc \rho}
  \end{pmatrix*}
\end{pmatrix*} \nonumber
\end{equation*}
Each \(\Fun{Stat}(X)\) is a category of stochastic channels with respect to measures on the space \(X\). We can write morphisms \(d^\dag : \Pow X \to \Kl(\Pow)(A, B)\) in \(\Fun{Stat}(X)\) as \(d^\dag_{(\cdot)} : A \xklto{(\cdot)} B\), and think of them  as generalized Bayesian inversions: given a measure \(\pi\) on \(X\), we obtain a channel \(d^\dag_\pi : A \xklto{\pi} B\) with respect to \(\pi\). Given a channel \(c : Y \klto X\) in the base category of priors, we can pull \(d^\dag\) back along \(c\), to obtain a \(Y\text{-dependent}\) channel in \(\Fun{Stat}(Y)\), \(c^\ast d^\dag : \Pow Y \to \Kl(\Pow)(A, B)\), which takes \(\rho : \Pow Y\) to the channel \(d^\dag_{c \klcirc \rho} : A \xklto{c \klcirc \rho} B\) defined by pushing \(\rho\) through \(c\) and then applying \(d^\dag\).
\end{ex}

\begin{rmk}
Note that by taking \(\Cat{Meas}\) to be Cartesian closed, we have \(\Cat{Meas}(\Pow X, \Cat{Meas}(A, \Pow B)) \cong \Cat{Meas}(\Pow X \times A, \Pow B)\) for each \(X\), \(A\) and \(B\), and so a morphism \(c^\dag : \Pow Y \to \Kl(\Pow)(X, Y)\) equivalently has the type \(\Pow Y \times X \to \Pow Y\). Paired with a channel \(c : Y \to \Pow X\), we have something like a Cartesian lens; and to compose such pairs, we can use the Grothendieck construction \citep{nLab2020Grothendieck,Spivak2019Generalized}.
\end{rmk}

\begin{defn}[$\Cat{GrLens}_{\Fun{Stat}}$] \label{def:stat-lens}
Instantiating the category of Grothendieck \(F\text{-lenses } \Cat{GrLens}_F\) (see \citep{Spivak2019Generalized})
with \(F = \Fun{Stat} : \cat{C}\op \to \Cat{V\mdash Cat}\), we obtain the category \(\Cat{GrLens}_\Fun{Stat}\) whose objects are pairs \((X, A)\) of objects of \(\cat{C}\) and whose morphisms \((X, A) \lensto (Y, B)\) are elements of the set
\begin{equation}
\Cat{GrLens}_\Fun{Stat} \big( (X, A), (Y, B) \big) \cong \cat{C}(X, Y) \times \Cat{V} \big( \cat{C}(I, X), \cat{C}(B, A) \big) \, .
\end{equation}
The identity \(\Fun{Stat}\text{-lens}\) on \((Y, A)\) is \((\id_Y, \id_A)\), where by abuse of notation \(\id_A : \cat{C}(I, Y) \to \cat{C}(A, A)\) is the constant map \(\id_A\) defined in \eqref{eq:stat} that takes any state on \(Y\) to the identity on \(A\). The sequential composite of \((c, c^\dag) : (X, A) \lensto (Y, B)\) and \((d, d^\dag) : (Y, B) \lensto (Z, C)\) is the \(\Fun{Stat}\text{-lens } \big( (d \klcirc c), (c^\dag \circ c^\ast d^\dag) \big) : (X, A) \lensto (Z, C)\) with \((d \klcirc c) : \cat{C}(X, Z)\) and where \((c^\dag \circ c^\ast d^\dag) : \Cat{V}\big(\cat{C}(I, X), \cat{C}(C, A)\big)\) takes a state \(\pi : \cat{C}(I, X)\) on \(X\) to the channel \(c^\dag_{\pi} \klcirc \d^\dag_{c \klcirc \pi}\). If we think of the notation \((\cdot)^\dag\) as denoting the operation of forming the Bayesian inverse of a channel (in the case where \(A = X\), \(B = Y\) and \(C = Z\)), then the main result of this section is to show that \((d \klcirc c)^\dag_\pi \overset{d \klcirc c \klcirc \pi}{\sim} c^\dag_{\pi} \klcirc \d^\dag_{c \klcirc \pi}\), where \(\overset{d \klcirc c \klcirc \pi}{\sim}\) denotes \((d \klcirc c \klcirc \pi)\text{-almost-equality}\) \citep[Definition 2.5]{Smithe2020Bayesian}.
\end{defn}

In order to give an optical form for \(\Cat{GrLens}_\Fun{Stat}\), we need to find two \(\cat{M}\text{-actegories}\) with a common category of actions \(\cat{M}\). Let \(\hat{\cat{C}}\) and \(\check{\cat{C}}\) denote the categories \(\hat{\cat{C}} := \Cat{V\mdash Cat}(\cat{C}\op, \Cat{V})\) and \(\check{\cat{C}} := \Cat{V\mdash Cat}(\cat{C}, \Cat{V})\) of presheaves and copresheaves on \(\cat{C}\), and consider the following natural isomorphisms.
\begin{align}
\Cat{GrLens}_\Fun{Stat} \big( (X, A), (Y, B) \big) & \cong \cat{C}(X, Y) \times \Cat{V} \big( \cat{C}(I, X), \cat{C}(B, A) \big) \nonumber \\
& \cong \int^{M \, : \, \cat{C}} \cat{C}(X, Y) \times \cat{C}(X, M) \times \Cat{V}\big(\cat{C}(I, M), \cat{C}(B, A)\big) \nonumber \\
& \cong \int^{\hat{M} \, : \, \hat{\cat{C}}} \cat{C}(X, Y) \times \hat{M}(X) \times \Cat{V}\big(\hat{M}(I), \cat{C}(B, A)\big) \label{eq:stat-lens-coend}
\end{align}
The second isomorphism follows by Yoneda reduction \citep{Loregian2015This,Roman2020Profunctor}, and the third follows by the Yoneda lemma. We take \(\cat{M}\) to be \(\cat{M} := \hat{\cat{C}}\), and define an action \(\odot\) of \(\hat{\cat{C}}\) on \(\check{\cat{C}}\) as follows.
\begin{defn}[$\odot$]
We give only the action on objects; the action on morphisms is analogous.
\begin{equation} \label{eq:L-action}
\begin{aligned}
\odot : \hat{\cat{C}} & \to \Cat{V\mdash Cat}(\check{\cat{C}}, \check{\cat{C}}) \\
\hat{M} & \mapsto
  \begin{pmatrix*}
    \hat{M} \odot - & : & \check{\cat{C}} & \to & \check{\cat{C}} \\
    & & P & \mapsto & \Cat{V}\big( \hat{M}(I), P \big)
  \end{pmatrix*}
\end{aligned}
\end{equation}
Functoriality of \(\odot\) follows from the functoriality of copresheaves. \qed
\end{defn}

\begin{prop} \label{prop:stat-actegory}
\(\odot\) equips \(\check{\cat{C}}\) with a \(\hat{\cat{C}}\text{-actegory}\) structure: unitor isomorphisms \(\lambda^{\odot}_F : 1 \odot F \xto{\sim} F\) and associator isomorphisms \(a^{\odot}_{\hat{M}, \hat{N}, F} : (\hat{M} \times \hat{N}) \odot F \xrightarrow{\sim} \hat{M} \odot (\hat{N} \odot F)\) for each \(\hat{M},\hat{N}\) in \(\check{\cat{C}}\), both natural in \(F : \Cat{V\mdash Cat}(\cat{C}, \Cat{V})\).
\end{prop}

We are now in a position to define the category of abstract Bayesian lenses, and show that this category coincides with the category of \(\Fun{Stat}\text{-lenses}\).
\begin{defn}[Bayesian lenses]
Denote by \(\Cat{BayesLens}\) the category of optics \(\Cat{Optic}_{\times, \odot}\) for the action of the Cartesian product on presheaf categories \(\times : \hat{\cat{C}} \to \Cat{V\mdash Cat}(\hat{\cat{C}}, \hat{\cat{C}})\) and the action \(\odot : \hat{\cat{C}} \to \Cat{V\mdash Cat}(\check{\cat{C}}, \check{\cat{C}})\) defined in \eqref{eq:L-action}. Its objects \((\hat{X}, \check{Y})\) are pairs of a presheaf and a copresheaf on \(\cat{C}\), and its morphisms \((\hat{X}, \check{A}) \lensto (\hat{Y}, \check{B})\) are abstract \emph{Bayesian lenses}---elements of the type
\begin{equation}
\Cat{Optic}_{\times, \odot}\Big((\hat{X}, \check{A}), (\hat{Y}, \check{B})\Big)
= \int^{\hat{M} \, : \, \hat{\cat{C}}} \hat{\cat{C}}(\hat{X}, \hat{M} \times \hat{Y}) \times \check{\cat{C}}(\hat{M} \odot \check{B}, \check{A})
\end{equation}
Given \(v : \cat{C}(X, Y)\) and \(u : \Cat{V}(\cat{C}(I, X), \cat{C}(B, A))\), we denote the corresponding element of this type by \(\optar{v}{u}\). A Bayesian lens \((\hat{X}, \check{X}) \lensto (\hat{Y}, \check{Y})\) is called a \textbf{simple} Bayesian lens.
\end{defn}

\begin{prop} \label{prop:bayeslens-are-lenses}
\(\Cat{BayesLens}\) is a category of lenses; a definition is given in \citep[§2.2.1]{Smithe2020Bayesian}.
\end{prop}

\begin{prop}[$\Fun{Stat}\text{-lenses}$ are Bayesian lenses] \label{prop:stat-lens-bayeslens}
Let \(\hat{(\cdot)} : \cat{C} \hookrightarrow \Cat{V\mdash Cat}(\cat{C}\op, \Cat{V})\) denote the Yoneda embedding and \(\check{(\cdot)} : \cat{C} \hookrightarrow \Cat{V\mdash Cat}(\cat{C}, \Cat{V})\) the coYoneda embedding. Then
\begin{equation}
\Cat{Optic}_{\times, \odot}\Big((\hat{X}, \check{A}), (\hat{Y}, \check{B})\Big)
\cong
\Cat{GrLens}_\Fun{Stat} \Big( (X, A), (Y, B) \Big)
\end{equation}
so that \(\Cat{GrLens}_\Fun{Stat}\) is equivalent to the full subcategory of \(\Cat{Optic}_{\times, \odot}\) on representable (co)presheaves.
\end{prop}

\begin{rmk}
We will often abuse notation by indicating representable objects in \(\Cat{BayesLens}\) by their representations in \(\cat{C}\). That is, we will write \((X, A)\) instead of \((\hat{X}, \check{A})\) where this would be unambiguous.
\end{rmk}

\begin{prop} \label{prop:bayeslens-smc}
\(\Cat{BayesLens}\) is a symmetric monoidal category. The monoidal product \(\otimes\) is inherited from \(\cat{C}\); the unit object is the pair \((I, I)\) where \(I\) is the unit object in \(\cat{C}\). For more details on the structure, see \citep{Riley2018Categories} or \citep{Moeller2018Monoidal}.
\end{prop}

\begin{defn}[Exact and approximate Bayesian lens]
Let \(\optar{c}{c^\dag} : (X, X) \lensto (Y, Y)\) be a simple Bayesian lens. We say that \(\optar{c}{c^\dag}\) is \textbf{exact} if \(c\) admits Bayesian inversion and, for each \(\pi : I \klto X\) such that \(c \klcirc \pi\) has non-empty support, \(c^\dag_\pi\) is the Bayesian inversion of \(c\) with respect to \(\pi\). Simple Bayesian lenses that are not exact are said to be \textbf{approximate}.
\end{defn}

\begin{lemma} \label{lemma:optical-bayes}
Let \(\optar{c}{c^\dag}\) and \(\optar{d}{d^\dag}\) be sequentially composable exact Bayesian lenses. Then the contravariant component of the composite lens \(\optar{d}{d^\dag} \lenscirc \optar{c}{c^\dag} \cong \optar{d \klcirc c}{c^\dag \circ c^\ast d^\dag}\) is, up to \(d \klcirc c \klcirc \pi \text{-almost-}\allowbreak\text{equality}\), the Bayesian inversion of \(d \klcirc c\) with respect to any state \(\pi\) on the domain of \(c\) such that \(c \klcirc \pi\) has non-empty support. That is to say, \emph{Bayesian updates compose optically}: \((d \klcirc c)^\dag_\pi \overset{d \klcirc c \klcirc \pi}{\sim} c^\dag_\pi \klcirc d^\dag_{c \klcirc \pi}\).
\end{lemma}

\section{Open Games for General Optics}
\label{sec:org675467e}
\label{sec:games}

In this section, we supply mild generalizations of the structures underlying open games, building on those in \citep{Bolt2019Bayesian}; at first, then, we consider games over arbitrary categories of optics \(\Cat{Optic}_{\circR, \circL}\). Subsequently, we use games over Bayesian lenses (in the category of optics \(\Cat{BayesLens}\) introduced above) to exemplify a number of canonical statistical concepts, such as maximum likelihood estimation and the variational autoencoder, and clarify their compositional structure using the notion of \emph{optimization game} (Definition \ref{def:opt-game}). Owing to space constraints, we omit most proofs in this section; they will appear in a full paper expanding the present abstract, and can be supplied at the request of the reader.

\begin{obs}
In the graphical calculus for the compact closed bicategory of profunctors \(\Cat{Prof}\) \citep{Roman2020Open}, the hom object \(\Cat{Optic}_{\circR, \circL}((X, A), (Y, B))\) has the depiction
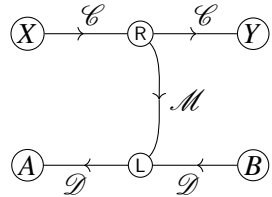
\[
\begin{tikzpicture}
	\begin{pgfonlayer}{nodelayer}
		\node [style=none] (9) at (0.5, 0) {};
		\node [style=none] (13) at (1.5, -1.75) {};
		\node [style=none] (14) at (1.5, 1.75) {};
		\node [style=none] (15) at (-1.5, -1.75) {};
		\node [style=none] (16) at (-1.5, 1.75) {};
		\node [style=dot] (17) at (-3, 1.75) {$X$};
		\node [style=dot] (18) at (-3, -1.75) {$A$};
		\node [style=action] (19) at (0, 1.75) {$\mathsf{R}$};
		\node [style=action] (20) at (0, -1.75) {$\mathsf{L}$};
		\node [style=dot] (21) at (3, 1.75) {$Y$};
		\node [style=dot] (22) at (3, -1.75) {$B$};
		\node [style=none] (23) at (-1.25, 2.25) {$\mathcal{C}$};
		\node [style=none] (24) at (1.25, 0) {$\mathcal{M}$};
		\node [style=none] (25) at (-1.75, -2.25) {$\mathcal{D}$};
		\node [style=none] (26) at (1.75, 2.25) {$\mathcal{C}$};
		\node [style=none] (27) at (1.25, -2.25) {$\mathcal{D}$};
	\end{pgfonlayer}
	\begin{pgfonlayer}{edgelayer}
		\draw [style=pointy1] (17) to (16.center);
		\draw (16.center) to (19);
		\draw [style=pointy1, in=90, out=-45, looseness=0.75] (19) to (9.center);
		\draw [in=45, out=-90, looseness=0.75] (9.center) to (20);
		\draw [style=pointy1] (20) to (15.center);
		\draw (15.center) to (18);
		\draw (13.center) to (20);
		\draw (14.center) to (21);
		\draw [style=pointy1] (19) to (14.center);
		\draw [style=pointy1] (22) to (13.center);
	\end{pgfonlayer}
\end{tikzpicture}
\]
where the types on the wires are the 0-cells of \(\Cat{Prof}\), the monoidal actions \(\circR\) and \(\circL\) are depicted as (co)monoids, and the states and effects are (co)representable functors on the objects \(X,A,Y,B\), treated as profunctors.
\end{obs}

\begin{defn}[Generalized context] \label{def:ctx}
The context functor \(C : \Cat{Optic}_{\circR, \circL}\op \times \Cat{Optic}_{\circR, \circL} \to \Set\) takes the pair of optical objects \(((X, A), (Y, B))\) to the type with depiction
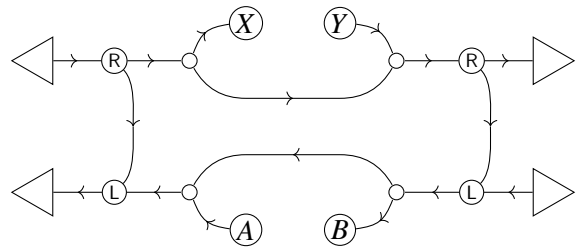
\[
\begin{tikzpicture}
	\begin{pgfonlayer}{nodelayer}
		\node [style=none] (0) at (1, 0) {};
		\node [style=state90] (5) at (-1.5, 1.75) {};
		\node [style=state90] (6) at (-1.5, -1.75) {};
		\node [style=action] (7) at (0.5, 1.75) {$\mathsf{R}$};
		\node [style=action] (8) at (0.5, -1.75) {$\mathsf{L}$};
		\node [style=copier] (9) at (2.5, 1.75) {};
		\node [style=copier] (10) at (2.5, -1.75) {};
		\node [style=dot] (11) at (4, 2.75) {$X$};
		\node [style=dot] (12) at (6.5, 2.75) {$Y$};
		\node [style=dot] (13) at (4, -2.75) {$A$};
		\node [style=dot] (14) at (6.5, -2.75) {$B$};
		\node [style=none] (15) at (10.5, 0) {};
		\node [style=copier] (20) at (8, 1.75) {};
		\node [style=copier] (21) at (8, -1.75) {};
		\node [style=action] (22) at (10, 1.75) {$\mathsf{R}$};
		\node [style=action] (23) at (10, -1.75) {$\mathsf{L}$};
		\node [style=state270] (24) at (12, 1.75) {};
		\node [style=state270] (25) at (12, -1.75) {};
		\node [style=none] (26) at (4, 0.75) {};
		\node [style=none] (27) at (6.5, 0.75) {};
		\node [style=none] (28) at (4, -0.75) {};
		\node [style=none] (29) at (6.5, -0.75) {};
	\end{pgfonlayer}
	\begin{pgfonlayer}{edgelayer}
		\draw [style=pointy1, in=90, out=-45, looseness=0.75] (7) to (0.center);
		\draw [in=45, out=-90, looseness=0.75] (0.center) to (8);
		\draw [style=pointy1, in=90, out=-45, looseness=0.75] (22) to (15.center);
		\draw [in=45, out=-90, looseness=0.75] (15.center) to (23);
		\draw [in=180, out=-60] (9) to (26.center);
		\draw [in=240, out=0] (27.center) to (20);
		\draw [in=360, out=120] (21) to (29.center);
		\draw [in=60, out=-180] (28.center) to (10);
		\draw [style=midpoint1, bend left, looseness=1.25] (21) to (14);
		\draw [style=midpoint1, bend left, looseness=1.25] (12) to (20);
		\draw [style=midpoint1, bend left, looseness=1.25] (9) to (11);
		\draw [style=midpoint1, bend left, looseness=1.25] (13) to (10);
		\draw [style=midpoint1] (20) to (22);
		\draw [style=midpoint1] (22) to (24);
		\draw [style=midpoint1] (25) to (23);
		\draw [style=midpoint1] (23) to (21);
		\draw [style=midpoint1] (5) to (7);
		\draw [style=midpoint1] (7) to (9);
		\draw [style=midpoint1] (10) to (8);
		\draw [style=midpoint1] (8) to (6);
		\draw [style=midpoint1] (26.center) to (27.center);
		\draw [style=midpoint1] (29.center) to (28.center);
	\end{pgfonlayer}
\end{tikzpicture}
\]
The triangles depict the (co)presheaves on the monoidal unit \(I\) in the underlying actegories. The action on morphisms (\emph{i.e.}, optics) is by precomposition on the left and postcomposition on the right. Functoriality follows accordingly. \qed

We can compose a context with an optic to obtain a `closed' system, as follows:
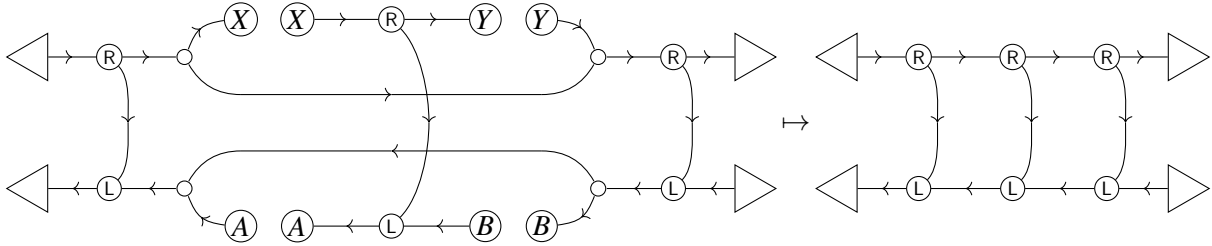
\[
\begin{tikzpicture}
	\begin{pgfonlayer}{nodelayer}
		\node [style=none] (0) at (-2, 0) {};
		\node [style=state90] (5) at (-4.5, 1.75) {};
		\node [style=state90] (6) at (-4.5, -1.75) {};
		\node [style=action] (7) at (-2.5, 1.75) {$\mathsf{R}$};
		\node [style=action] (8) at (-2.5, -1.75) {$\mathsf{L}$};
		\node [style=copier] (9) at (-0.5, 1.75) {};
		\node [style=copier] (10) at (-0.5, -1.75) {};
		\node [style=dot] (11) at (1, 2.75) {$X$};
		\node [style=dot] (12) at (9, 2.75) {$Y$};
		\node [style=dot] (13) at (1, -2.75) {$A$};
		\node [style=dot] (14) at (9, -2.75) {$B$};
		\node [style=none] (15) at (13, 0) {};
		\node [style=copier] (20) at (10.5, 1.75) {};
		\node [style=copier] (21) at (10.5, -1.75) {};
		\node [style=action] (22) at (12.5, 1.75) {$\mathsf{R}$};
		\node [style=action] (23) at (12.5, -1.75) {$\mathsf{L}$};
		\node [style=state270] (24) at (14.5, 1.75) {};
		\node [style=state270] (25) at (14.5, -1.75) {};
		\node [style=none] (26) at (1, 0.75) {};
		\node [style=none] (27) at (9, 0.75) {};
		\node [style=none] (28) at (1, -0.75) {};
		\node [style=none] (29) at (9, -0.75) {};
		\node [style=none] (30) at (5, 0.75) {};
		\node [style=none] (31) at (5, -0.75) {};
		\node [style=none] (32) at (6, 0) {};
		\node [style=dot] (37) at (2.5, 2.75) {$X$};
		\node [style=dot] (38) at (2.5, -2.75) {$A$};
		\node [style=action] (39) at (5, 2.75) {$\mathsf{R}$};
		\node [style=action] (40) at (5, -2.75) {$\mathsf{L}$};
		\node [style=dot] (41) at (7.5, 2.75) {$Y$};
		\node [style=dot] (42) at (7.5, -2.75) {$B$};
		\node [style=none] (43) at (6.25, 0.75) {};
	\end{pgfonlayer}
	\begin{pgfonlayer}{edgelayer}
		\draw [style=pointy1, in=90, out=-45, looseness=0.75] (7) to (0.center);
		\draw [in=45, out=-90, looseness=0.75] (0.center) to (8);
		\draw [style=pointy1, in=90, out=-45, looseness=0.75] (22) to (15.center);
		\draw [in=45, out=-90, looseness=0.75] (15.center) to (23);
		\draw [style=pointy1] (26.center) to (30.center);
		\draw [style=pointy1] (29.center) to (31.center);
		\draw [in=180, out=-60] (9) to (26.center);
		\draw (30.center) to (27.center);
		\draw [in=240, out=0] (27.center) to (20);
		\draw [in=360, out=120] (21) to (29.center);
		\draw (31.center) to (28.center);
		\draw [in=60, out=-180] (28.center) to (10);
		\draw [style=midpoint1, bend left, looseness=1.25] (21) to (14);
		\draw [style=midpoint1, bend left, looseness=1.25] (12) to (20);
		\draw [style=midpoint1, bend left, looseness=1.25] (9) to (11);
		\draw [style=midpoint1, bend left, looseness=1.25] (13) to (10);
		\draw [style=midpoint1] (20) to (22);
		\draw [style=midpoint1] (22) to (24);
		\draw [style=midpoint1] (25) to (23);
		\draw [style=midpoint1] (23) to (21);
		\draw [style=midpoint1] (5) to (7);
		\draw [style=midpoint1] (7) to (9);
		\draw [style=midpoint1] (10) to (8);
		\draw [style=midpoint1] (8) to (6);
		\draw [style=pointy1, in=90, out=-45, looseness=0.75] (39) to (32.center);
		\draw [in=45, out=-90, looseness=0.75] (32.center) to (40);
		\draw [style=midpoint1] (37) to (39);
		\draw [style=midpoint1] (39) to (41);
		\draw [style=midpoint1] (42) to (40);
		\draw [style=midpoint1] (40) to (38);
	\end{pgfonlayer}
\end{tikzpicture} \mapsto 
\begin{tikzpicture}
	\begin{pgfonlayer}{nodelayer}
		\node [style=none] (0) at (-2, 0) {};
		\node [style=state90] (5) at (-4.5, 1.75) {};
		\node [style=state90] (6) at (-4.5, -1.75) {};
		\node [style=action] (7) at (-2.5, 1.75) {$\mathsf{R}$};
		\node [style=action] (8) at (-2.5, -1.75) {$\mathsf{L}$};
		\node [style=none] (15) at (3, 0) {};
		\node [style=action] (22) at (2.5, 1.75) {$\mathsf{R}$};
		\node [style=action] (23) at (2.5, -1.75) {$\mathsf{L}$};
		\node [style=state270] (24) at (4.5, 1.75) {};
		\node [style=state270] (25) at (4.5, -1.75) {};
		\node [style=none] (26) at (0.5, 0) {};
		\node [style=action] (27) at (0, 1.75) {$\mathsf{R}$};
		\node [style=action] (28) at (0, -1.75) {$\mathsf{L}$};
	\end{pgfonlayer}
	\begin{pgfonlayer}{edgelayer}
		\draw [style=pointy1, in=90, out=-45, looseness=0.75] (7) to (0.center);
		\draw [in=45, out=-90, looseness=0.75] (0.center) to (8);
		\draw [style=pointy1, in=90, out=-45, looseness=0.75] (22) to (15.center);
		\draw [in=45, out=-90, looseness=0.75] (15.center) to (23);
		\draw [style=midpoint1] (22) to (24);
		\draw [style=midpoint1] (25) to (23);
		\draw [style=midpoint1] (5) to (7);
		\draw [style=midpoint1] (8) to (6);
		\draw [style=pointy1, in=90, out=-45, looseness=0.75] (27) to (26.center);
		\draw [in=45, out=-90, looseness=0.75] (26.center) to (28);
		\draw [style=midpoint1] (7) to (27);
		\draw [style=midpoint1] (27) to (22);
		\draw [style=midpoint1] (23) to (28);
		\draw [style=midpoint1] (28) to (8);
	\end{pgfonlayer}
\end{tikzpicture}
\]
\end{defn}

\begin{conjecture} \label{conj:doubling}
It is easy to show that a context on \(((X,A),(Y,B))\) is equivalently a state \((I, I) \lensto ((X,A),(Y,B))\) in the monoidal category of `double lenses', \(\Cat{Lens}_{\Cat{Optic}_{\circR, \circL}}\) \citep{Bolt2019Bayesian}. Rendering this graphically leads us to the following conjecture: categories of double optics are instances of the \emph{doubling} or \emph{CP} construction from categorical quantum mechanics (\emph{cf.} \citep{Coecke2015Categorical,Coecke2016Categorical}).
\end{conjecture}

\begin{prop} \label{prop:ctx-nice}
Let \(\cat{C}\) and \(\cat{D}\) be the (monoidal) actegories underlying \(\Cat{Optic}_{\circR, \circL}\), and denote their respective monoidal units by \(I_{\cat{C}}\) and \(I_{\cat{D}}\). If these unit objects are terminal in their respective categories, then the contexts \(C((X, A), (Y, B))\) simplify to
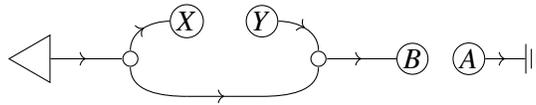
\[
\begin{tikzpicture}
	\begin{pgfonlayer}{nodelayer}
		\node [style=copier] (0) at (-2.5, 0) {};
		\node [style=copier] (1) at (2.5, 0) {};
		\node [style=dot] (2) at (-1, 1) {$X$};
		\node [style=dot] (3) at (1, 1) {$Y$};
		\node [style=none] (4) at (-1, -1) {};
		\node [style=none] (5) at (1, -1) {};
		\node [style=state90] (6) at (-5, 0) {};
		\node [style=dot] (7) at (5, 0) {$B$};
		\node [style=dot] (8) at (6.5, 0) {$A$};
		\node [style=discarder0, rotate=0] (9) at (8, 0) {};
	\end{pgfonlayer}
	\begin{pgfonlayer}{edgelayer}
		\draw [style=midpoint1] (6) to (0);
		\draw [style=midpoint1, in=-180, out=90] (0) to (2);
		\draw [style=midpoint1] (4.center) to (5.center);
		\draw [style=midpoint1, in=90, out=0] (3) to (1);
		\draw [style=midpoint1] (1) to (7);
		\draw [in=-180, out=-90] (0) to (4.center);
		\draw [in=-90, out=0] (5.center) to (1);
		\draw [style=midpoint1] (8) to (9);
	\end{pgfonlayer}
\end{tikzpicture}
\]
where we have depicted the representable presheaf on \(I_{\cat{D}}\) as \(\ground\) to indicate that \(A\) is just discarded. Consequently, in this case, a context is just an optic \((I, B) \lensto (X, Y)\).

\end{prop}

\begin{defn}[Generalized open game] \label{def:open-game}
Let \((X, A)\) and \((Y, B)\) be objects in any symmetric monoidal category of optics \(\Cat{Optic}_{\circR, \circL}\). Let \(\Sigma\) be a \(\cat{U}\text{-category}\), for any base of enrichment \(\cat{U}\) such that \(\cat{U}\text{-}\Cat{Prof}\) is compact closed. An \textbf{open game} from \((X, A)\) to \((Y, B)\) with strategies in \(\Sigma\), denoted \(G : (X, A) \xto{\Sigma} (Y, B)\), is given by:
\begin{enumerate}
\item a play function \(P : \Sigma_0 \to \Cat{Optic}_{\circR, \circL}((X, A), (Y, B))\); and
\item a best response function \(B : C((X, A), (Y, B)) \to \cat{U}\text{-}\Cat{Prof}(\Sigma, \Sigma)\).
\end{enumerate}
Given a strategy \(\sigma : \Sigma\), we will often write \(\optar{v}{u}_\sigma\) or similar to denote its image under \(P\). A strategy is an \textbf{equilibrium} in a context \(\optar{\pi}{k}\) if it is a fixed point of \(B(\optar{\pi}{k})\).
\end{defn}

Roughly speaking, the `best responses' to a strategy \(\sigma\) in a context is are those strategies \(\tau\) such that choosing \(\tau\) would result in performance at the game at least as good as choosing \(\sigma\); equilibrium strategies are those for which such deviation would not improve performance.

\begin{rmk} \label{rmk:b-relt}

Note: whereas classic open games use a best-response relation, we categorify that here to a best-response \emph{relator} (in the terminology of \citep{Loregian2015This}; \emph{i.e.}, a `proof-relevant' relation), so that we can describe the trajectories witnessing the computation of equilibria, rather than their mere existence.
\end{rmk}

\begin{prop} \label{prop:cat-open-games}
Generalized open games over the symmetric monoidal category of optics \(\Cat{Optic}_{\circR, \circL}\) with strategies enriched in \(\cat{U}\) form a symmetric monoidal category denoted \(\Cat{Game}(\cat{U}, \circR, \circL)\).
\end{prop}

Since our games are only a mild generalization of those of \citep{Bolt2019Bayesian}, we refer the reader to §3.10 of that paper for an idea of the proof of the foregoing proposition, which goes through analogously. The sequential composition of games is given by the sequential composition of optics, with the best response to the composite being the product of the best responses to the factors. Similarly, parallel composition is given by the monoidal product of optics, and the best response to the composite is again the product of the best responses to the factors.

We now consider some games over \(\Cat{BayesLens}\) that supply the building blocks of the archetypal cybernetic systems to be considered in \secref{sec:cyber-sys}. For now, we will take the strategies simply to be discrete categories (\emph{i.e.}, sets), as in the standard formulation of open games. Consequently, we will take the codomain of the best response function to be \(\Set(\Sigma, \Set(\Sigma, 2))\), for each strategy type \(\Sigma\). We assume the ambient category of stochastic channels is semicartesian, so that the monoidal unit is the terminal object.

\begin{rmk}
All the games we will consider henceforth will have play functions whose codomains restrict to the representable subcategory \(\Cat{GrLens}_\Fun{Stat}\) of \(\Cat{BayesLens}\); in this work, we do not use the extra generality afforded by \(\Cat{BayesLens}\), except insofar as it grants us the use of string diagrams in \(\Cat{Prof}\), which we find helpful for reasoning intuitively about these systems. The generality of optics \emph{is} however used in the `game-theoretic' games of \citep{Bolt2019Bayesian}, and in future work we hope to relate the cybernetic systems of this paper to the game-theoretic setting of that earlier work.
\end{rmk}

\begin{rmk} \label{rmk:atomic-games}
All the statistical games considered in this paper will be `atomic' in the sense of \citep{Bolt2019Bayesian}: in particular, the best response functions we consider will be constant, meaning that, in any context, the set of best strategies does not depend on the `current' choice of strategy. Permitting such dependence will be important in future work, however, when we consider how cybernetic systems interact, and hence respond to each other.
\end{rmk}

\begin{ex} \label{ex:ml-game}
A Bayesian lens of the form \((I, I) \lensto (X, X)\) is fully specified by a state \(\pi : I \klto X\). A context for such a lens is given by a lens \(\optar{!}{k} : (I, X) \lensto (X, X)\) where \(! : I \klto I\) is the unique map and \(k : X \klto X\) is any endochannel on \(X\). A \textbf{maximum likelihood game} is any game whose play function has codomain in Bayesian lenses of this form \((I, I) \lensto (X, X)\) for any \(X : \cat{C}\), and whose best response function is isomorphic to
\[
B(\optar{!}{k}) = \optar{\rho}{!}_\sigma \mapsto \left\{ \optar{\pi}{!}_\tau \middle| \pi \in \underset{\pi : I \klto X}{\arg\max} \E_{k \klcirc \pi} \left[ \pi \right]  \right\}
\]
where \(\E\) is the canonical expectation operator (\emph{i.e.} algebra evaluation) associated to states in \(\cat{C}\), and where we have written \(\optar{\rho}{!}_\sigma\) and \(\optar{\pi}{!}_\tau\) to denote the images of the strategies \(\sigma\) and \(\tau\) under the play function. Intuitively, then, the best response is given by the strategy that maximises the likelihood of the state obtained from the context \(k\).
\end{ex}

\begin{rmk}
In what follows, we assume that the underlying category \(\cat{C}\) of stochastic channels \emph{admits density functions}. Informally, a density function for a stochastic channel \(c : X \klto Y\) is a measurable function \(p_c : Y \times X \to [0, 1]\) whose values are the probabilities (or probability densities) \(p_c(y | x)\)  at each pair \((y, x) : Y \times X\). We say that the value \(p_c(y | x)\) is the probability (or probability density) of \(y\) \emph{given} \(x\). In a category such as \(\Kl(\Dst_{\leq 1})\), whose objects are sets and whose morphisms \(X \klto Y\) are functions \(X \to \Dst(Y + 1)\), a density function for \(c : X \klto Y\) is a morphism \(Y \otimes X \klto I\); note that in \(\Kl(\Dst_{\leq 1})\), \(I\) is not terminal. In the finitely-supported case, density functions are effectively equivalent to channels, but this is not the case in the continuous setting, where they are of most use. For more on this, see \citep[§2.1.4]{Smithe2020Bayesian}.
\end{rmk}

A natural first generalization of maximum likelihood games takes us from states \(I \klto X\) to channels \(Z \klto X\); that is, from `elements' to `generalized elements' in the covariant (forwards) part of the lens. Unlike Bayesian lenses \((I, I) \lensto (X, X)\), lenses \((Z, Z) \lensto (X, X)\) admit nontrivial contravariant components, which we think of as generalized Bayesian inversions. Consequently, our first generalization is a notion of `Bayesian inference game'. A context \(\optar{\pi}{k} : (I, X) \lensto (Z, X)\) for a Bayesian lens \((Z, Z) \lensto (X, X)\) then constitutes a `prior' state \(\pi : I \klto Z\) and a `continuation' channel \(k : X \klto X\) which together witness the closure of the otherwise open system.

\begin{ex} \label{ex:simp-inf-game}
Fix a channel \(c : Z \klto X\) with associated density function \(p_c : X \times Z \to \rr_+\) and a measure of divergence between states on \(Z\), \(D : \cat{C}(I, Z) \times \cat{C}(I, Z) \to \rr\). A corresponding (generalized) \textbf{simple Bayesian inference game} is any game whose play function has codomain \(\Cat{BayesLens}((Z, Z), (X, X))\) and whose best response function is isomorphic to
\begin{align*}
B(\optar{\pi}{k}) = \optar{d}{d'}_\sigma & \mapsto
\Bigg\{ \optar{c}{c'}_\tau \bigg| 
c' \in \underset{c' : \Cat{V}(\cat{C}(I, Z), \, \cat{C}(X, Z))}{\arg\min} \E_{x \sim k \klcirc c \klcirc \pi} \bigg[ \E_{z \sim c'_\pi(x)} \left[ - \log p_c(x | z)  \right]
  + D(c'_\pi(x), \pi) \bigg] \Bigg\} \\
= \optar{d}{d'}_\sigma & \mapsto
\Bigg\{ \optar{c}{c'}_\tau \bigg| 
c' \in \underset{c' : \Cat{V}(\cat{C}(I, Z), \, \cat{C}(X, Z))}{\arg\min} \bigg( \E_{z \sim c'_\pi \klcirc k \klcirc c \klcirc \pi} \left[ - \int_X \log p_c(\d k \klcirc c \klcirc \pi | z)  \right] \\
& \qquad\qquad\qquad\qquad\qquad\qquad\qquad\qquad\qquad
  + D(c'_\pi \klcirc k \klcirc c \klcirc \pi, \pi) \bigg) \Bigg\}
\end{align*}
where \(\pi : I \klto Z\) and \(k : X \klto X\), and where the notation \(z \sim \pi\) means ``\(z\) distributed according to the state \(\pi\)''. Note that the second line follows from the first by linearity of expectation.
\end{ex}

\begin{prop}[{\citep[Thm. 1]{Knoblauch2019Generalized}}]
When \(D\) is chosen to be the Kullback-Leibler divergence \(D_{KL}\), minimizing the objective function defining a simple Bayesian inference game is equivalent to computing an (exact) Bayesian inversion.
\end{prop}

\begin{cor}
Given two Bayesian inference games \(G : (Z, Z) \lensto (Y, Y)\) and \(H : (Y, Y) \lensto (X, X)\), we can compose them sequentially to obtain a game \(H \lenscirc G : (Z, Z) \lensto (X, X)\), which we will call a \textbf{hierarchical Bayesian inference game}. It is then an immediate consequence of Lemma \ref{lemma:optical-bayes} that, in any given context for which the forwards channels admit Bayesian inversion, the best response to the composite game \(H \lenscirc G\) (that is, the optimal inversion of the composite channel) is given simply by (the composition of) the best responses to the factors \(H\) and \(G\). Consequently, Bayesian inference games are closed under composition.
\end{cor}

Similarly, given a channel \(c : Z \otimes Y \klto X\), we can consider the \textbf{marginal Bayesian inference game} in which the objective is to compute the inversion of the channel onto just one of the factors \(Z\) or \(Y\) in the domain.

\begin{ex}[Variational autoencoder game] \label{ex:vae-game}
Fix a family \(\mathcal{F} \hookrightarrow \cat{C}(Z, X)\) of forward channels and a family \(\mathcal{P} \hookrightarrow \cat{C}(X, Z)\) of backward channels such that each \(c : \mathcal{F}\) admits a density function \(p_c : X \otimes Z \to \rr_+\) and each \(d : \mathcal{P}\) admits a density function \(q : Z \otimes X \to \rr_+\); think of these families as determining parameterizations of the channels. We take our strategy type to be \(\Sigma = \mathcal{F} \times \mathcal{P}\). A \textbf{simple variational autoencoder game} \((Z,Z) \xto{\Sigma} (X,X)\) is any game with play function \(P : \Sigma \to \Cat{BayesLens}((Z,Z), (X,X))\) and whose best response function is isomorphic to
\begin{align*}
B(\optar{\pi}{k}) = \optar{d}{d'}_\sigma \mapsto
\Bigg\{ \optar{c}{c'}_\tau \Bigg| (c, c') \in \underset{\substack{c \in \mathcal{F}, \\ c' \in \Cat{V}(\cat{C}(I, Z), \, \mathcal{P})}}{\arg\min} \E_{x \sim k \klcirc c \klcirc \pi} \E_{z \sim c'_\pi(x)} \left[ \log \frac{q(z|x)}{p_c(x|z)p_\pi(z)} \right]
\Bigg\}
\end{align*}
where \(\pi : I \klto Z\) admits a density function \(p_\pi : Z \to \rr_+\), \(q : Z \otimes X \to \rr_+\) is a density function associated to \(c'_\pi\), and \(k\) has type \(X \klto X\).
\end{ex}

\begin{prop} \label{prop:vae-learns-model}
A best response to a variational autoencoder game is a stochastic channel \(c : \mathcal{F}\) that maximises the likelihood of the state observed through the continuation \(k\) under the assumption that the generative process is in \(\mathcal{F}\), along with an inverse channel \(c'_\pi : \mathcal{P}\) that best approximates the exact Bayesian inverse \(c^\dag_\pi\) under the constraint of being in \(\mathcal{P}\).
\end{prop}

\begin{prop} \label{prop:vae-infers}
Variational autoencoder games generalize inference games for the Kullback-Leibler divergence. More precisely, the objective function defining autoencoder games is of the same form as that defining inference games \eqref{ex:simp-inf-game} when \(D = D_{KL}\).
\end{prop}

This prompts the following generalization:

\begin{ex}[Generalized autoencoder game] \label{ex:ae-game}
Fix two families of channels \(\mathcal{F},\mathcal{P}\) and a strategy type \(\Sigma\) as in Example \ref{ex:vae-game}. Then a (generalized) \textbf{simple autoencoder game} \((Z,Z) \xto{\Sigma} (X,X)\) is any game with play function \(P : \Sigma \to \Cat{BayesLens}((Z,Z), (X,X))\) and whose best response function is isomorphic to
\begin{align*}
B(\optar{\pi}{k}) = \optar{d}{d'}_\sigma \mapsto
\Bigg\{ \optar{c}{c'}_\tau \Bigg| (c, c') \in \underset{\substack{c \in \mathcal{F}, \\ c' \in \Cat{V}(\cat{C}(I, Z), \, \mathcal{P})}}{\arg\min} \bigg( & \E_{z \sim c'_\pi \klcirc k \klcirc c \klcirc \pi} \left[ - \int_X \log p_c(\d k \klcirc c \klcirc \pi | z) \right] \\
& \qquad + D(c'_\pi \klcirc k \klcirc c \klcirc \pi, \pi) \bigg)
\Bigg\}
\end{align*}
where \(\pi\) and \(k\) have respective types \(I \klto Z\) and \(X \klto X\), and \(D\) is any measure of divergence between states.

As with Bayesian inference games, we can generalize simple autoencoder games to \textbf{hierarchical} and \textbf{marginal} autoencoder games via the corresponding sequential and parallel compositions.
\end{ex}

The foregoing games have been purely statistically formulated, without capturing the motivating feature of an open system as something in interaction with an external environment. Nonetheless, we can model a simple open system of hierarchical \textbf{active inference} that receives stochastic inputs from an environment and emits actions stochastically into the environment, as follows.
\begin{ex}[Active inference game] \label{ex:ai-game}
Let \(\{S_i\}_i\) be set of spaces of sensory data indexed by hierarchical levels of abstraction \(i\) (for instance, the levels of abstraction might range from representations of whole objects to fine details about their texture); similarly, let \(\{A_i\}_i\) be a set of spaces of possible actions similarly hierarchically organized. Consider the marginal autoencoder games \((S_{i+1} \otimes A_{i}, S_{i+1}) \lensto (S_{i+1}, S_{i+1})\) and \((A_{i+1} \otimes S_{i}, A_{i+1}) \lensto (A_{i+1}, A_{i+1})\) coupled via the symmetric monoidal structure \(\otimes\) of \(\cat{C}\):
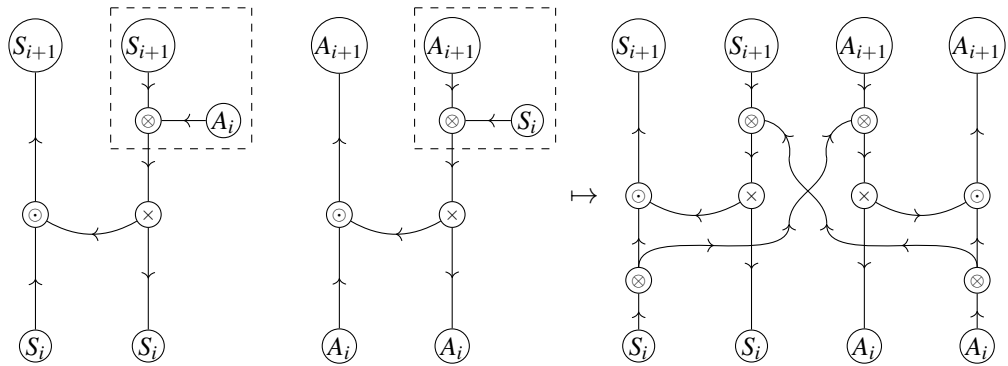
\[
\begin{tikzpicture}
	\begin{pgfonlayer}{nodelayer}
		\node [style=dot] (17) at (1.5, 4) {{\small $S_{i+1}$}};
		\node [style=dot] (18) at (-1.5, 4) {{\small $S_{i+1}$}};
		\node [style=action] (19) at (1.5, -0.5) {$\times$};
		\node [style=action] (20) at (-1.5, -0.5) {$\odot$};
		\node [style=dot] (21) at (1.5, -4) {{\small $S_{i}$}};
		\node [style=dot] (22) at (-1.5, -4) {{\small $S_{i}$}};
		\node [style=action] (23) at (1.5, 2) {$\otimes$};
		\node [style=dot] (24) at (3.5, 2) {{\small $A_{i}$}};
		\node [style=none] (25) at (0.5, 5) {};
		\node [style=none] (26) at (4.25, 5) {};
		\node [style=none] (27) at (0.5, 1.25) {};
		\node [style=none] (28) at (4.25, 1.25) {};
	\end{pgfonlayer}
	\begin{pgfonlayer}{edgelayer}
		\draw [style=midpoint1, bend left] (19) to (20);
		\draw [style=midpoint1] (20) to (18);
		\draw [style=midpoint1] (17) to (23);
		\draw [style=midpoint1] (24) to (23);
		\draw [style=midpoint1] (23) to (19);
		\draw [style=midpoint1] (19) to (21);
		\draw [style=midpoint1] (22) to (20);
		\draw [style=dashed1] (25.center) to (26.center);
		\draw [style=dashed1] (26.center) to (28.center);
		\draw [style=dashed1] (25.center) to (27.center);
		\draw [style=dashed1] (27.center) to (28.center);
	\end{pgfonlayer}
\end{tikzpicture} \qquad
\begin{tikzpicture}
	\begin{pgfonlayer}{nodelayer}
		\node [style=dot] (17) at (1.5, 4) {{\small $A_{i+1}$}};
		\node [style=dot] (18) at (-1.5, 4) {{\small $A_{i+1}$}};
		\node [style=action] (19) at (1.5, -0.5) {$\times$};
		\node [style=action] (20) at (-1.5, -0.5) {$\odot$};
		\node [style=dot] (21) at (1.5, -4) {{\small $A_{i}$}};
		\node [style=dot] (22) at (-1.5, -4) {{\small $A_{i}$}};
		\node [style=action] (23) at (1.5, 2) {$\otimes$};
		\node [style=dot] (24) at (3.5, 2) {{\small $S_{i}$}};
		\node [style=none] (25) at (0.5, 5) {};
		\node [style=none] (26) at (4.25, 5) {};
		\node [style=none] (27) at (0.5, 1.25) {};
		\node [style=none] (28) at (4.25, 1.25) {};
	\end{pgfonlayer}
	\begin{pgfonlayer}{edgelayer}
		\draw [style=midpoint1, bend left] (19) to (20);
		\draw [style=midpoint1] (20) to (18);
		\draw [style=midpoint1] (17) to (23);
		\draw [style=midpoint1] (24) to (23);
		\draw [style=midpoint1] (23) to (19);
		\draw [style=midpoint1] (19) to (21);
		\draw [style=midpoint1] (22) to (20);
		\draw [style=dashed1] (25.center) to (26.center);
		\draw [style=dashed1] (26.center) to (28.center);
		\draw [style=dashed1] (25.center) to (27.center);
		\draw [style=dashed1, in=180, out=0] (27.center) to (28.center);
	\end{pgfonlayer}
\end{tikzpicture} \; \mapsto \;
\begin{tikzpicture}
	\begin{pgfonlayer}{nodelayer}
		\node [style=dot] (0) at (-1.5, 4) {{\small $S_{i+1}$}};
		\node [style=dot] (1) at (-4.5, 4) {{\small $S_{i+1}$}};
		\node [style=action] (2) at (-1.5, 0) {$\times$};
		\node [style=action] (3) at (-4.5, 0) {$\odot$};
		\node [style=dot] (4) at (-1.5, -4) {{\small $S_{i}$}};
		\node [style=dot] (5) at (-4.5, -4) {{\small $S_{i}$}};
		\node [style=action] (6) at (-1.5, 2) {$\otimes$};
		\node [style=dot] (8) at (1.5, 4) {{\small $A_{i+1}$}};
		\node [style=dot] (9) at (4.5, 4) {{\small $A_{i+1}$}};
		\node [style=action] (10) at (1.5, 0) {$\times$};
		\node [style=action] (11) at (4.5, 0) {$\odot$};
		\node [style=dot] (12) at (1.5, -4) {{\small $A_{i}$}};
		\node [style=dot] (13) at (4.5, -4) {{\small $A_{i}$}};
		\node [style=action] (14) at (1.5, 2) {$\otimes$};
		\node [style=none] (16) at (-0.5, 1) {};
		\node [style=none] (17) at (0.5, -0.75) {};
		\node [style=none] (18) at (0.5, 1) {};
		\node [style=none] (19) at (-0.5, -0.75) {};
		\node [style=action] (34) at (-4.5, -2.25) {$\otimes$};
		\node [style=action] (35) at (4.5, -2.25) {$\otimes$};
	\end{pgfonlayer}
	\begin{pgfonlayer}{edgelayer}
		\draw [style=midpoint1, bend left] (2) to (3);
		\draw [style=midpoint1] (3) to (1);
		\draw [style=midpoint1] (0) to (6);
		\draw [style=midpoint1] (6) to (2);
		\draw [style=midpoint1, bend right] (10) to (11);
		\draw [style=midpoint1] (11) to (9);
		\draw [style=midpoint1] (8) to (14);
		\draw [style=midpoint1] (14) to (10);
		\draw [in=-90, out=90] (17.center) to (16.center);
		\draw [in=-90, out=90] (19.center) to (18.center);
		\draw [style=midpoint1, bend left=45] (18.center) to (14);
		\draw [style=midpoint1, bend right=45] (16.center) to (6);
		\draw [style=midpointy1pointy, in=-90, out=90, looseness=0.75] (34) to (19.center);
		\draw [style=midpointy1pointy, in=-90, out=90, looseness=0.75] (35) to (17.center);
		\draw [style=midpoint1] (5) to (34);
		\draw [style=midpoint1] (34) to (3);
		\draw [style=midpoint1] (2) to (4);
		\draw [style=midpoint1] (10) to (12);
		\draw [style=midpoint1] (13) to (35);
		\draw [style=midpoint1] (35) to (11);
	\end{pgfonlayer}
\end{tikzpicture}
\]
giving a composite game \((S_{i+1} \otimes A_{i+1}, S_{i+1} \times A_{i+1}) \lensto (S_{i} \otimes A_{i}, S_{i} \times A_{i})\). Recall from \citep[§§3.7-3.8]{Bolt2019Bayesian} that a composite game is given by the (sequential and parallel) composition of optics, with best-response given by the product of the best-responses of the factors.

Note that the Bayesian posterior inferred by such a game has independent factors on \(S_{i+1}\) and \(A_{i+1}\). This is not merely a diagrammatic convenience, but coincides with a common `mean field' simplification in the modelling literature \citep{Buckley2017free,Kingma2017Variational}. The dashed box is a functorial box \citep{Mellies2006Functorial} depicting the Yoneda embedding; recall that optics in \(\Cat{BayesLens}\) were defined over (co)presheaves, and so here we needed to lift the monoidal product on \(\cat{C}\) into a diagram over its presheaf category \(\Cat{Cat}(\cat{C}\op, \Set)\).

Next, compose these games along the hierarchy indexed by \(i\), to obtain a game \((S_{N} \otimes A_{N}, S_{N} \times A_{N}) \lensto (S_{0} \otimes A_{0}, S_{0} \times A_{0})\), such as an element of the following object:
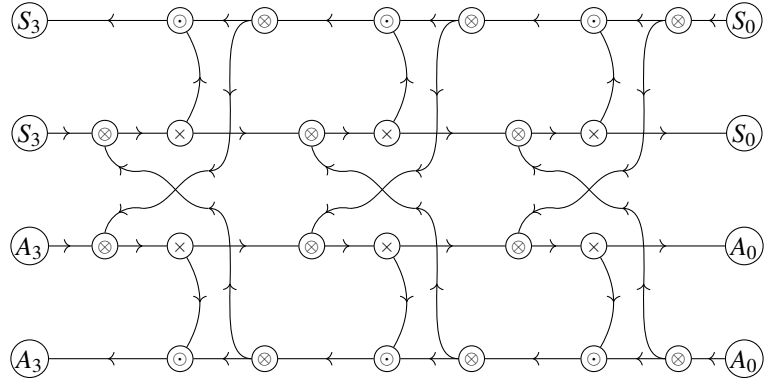
\[
\begin{tikzpicture}
	\begin{pgfonlayer}{nodelayer}
		\node [style=dot] (0) at (-9.5, 1.5) {{\small $S_{3}$}};
		\node [style=dot] (1) at (-9.5, 4.5) {{\small $S_{3}$}};
		\node [style=action] (2) at (-5.5, 1.5) {$\times$};
		\node [style=action] (3) at (-5.5, 4.5) {$\odot$};
		\node [style=action] (6) at (-7.5, 1.5) {$\otimes$};
		\node [style=dot] (8) at (-9.5, -1.5) {{\small $A_{3}$}};
		\node [style=dot] (9) at (-9.5, -4.5) {{\small $A_{3}$}};
		\node [style=action] (10) at (-5.5, -1.5) {$\times$};
		\node [style=action] (11) at (-5.5, -4.5) {$\odot$};
		\node [style=action] (14) at (-7.5, -1.5) {$\otimes$};
		\node [style=none] (16) at (-6.5, 0.5) {};
		\node [style=none] (18) at (-6.5, -0.5) {};
		\node [style=action] (36) at (0, 1.5) {$\times$};
		\node [style=action] (37) at (0, 4.5) {$\odot$};
		\node [style=action] (43) at (0, -1.5) {$\times$};
		\node [style=action] (44) at (0, -4.5) {$\odot$};
		\node [style=action] (68) at (5.5, 1.5) {$\times$};
		\node [style=action] (69) at (5.5, 4.5) {$\odot$};
		\node [style=dot] (70) at (9.5, 1.5) {{\small $S_{0}$}};
		\node [style=dot] (71) at (9.5, 4.5) {{\small $S_{0}$}};
		\node [style=action] (75) at (5.5, -1.5) {$\times$};
		\node [style=action] (76) at (5.5, -4.5) {$\odot$};
		\node [style=dot] (77) at (9.5, -1.5) {{\small $A_{0}$}};
		\node [style=dot] (78) at (9.5, -4.5) {{\small $A_{0}$}};
		\node [style=none] (98) at (-4.75, 0.5) {};
		\node [style=none] (99) at (-4.75, -0.5) {};
		\node [style=action] (100) at (-3.25, -4.5) {$\otimes$};
		\node [style=action] (101) at (-3.25, 4.5) {$\otimes$};
		\node [style=none] (102) at (0.75, 0.5) {};
		\node [style=none] (103) at (0.75, -0.5) {};
		\node [style=action] (104) at (2.25, -4.5) {$\otimes$};
		\node [style=action] (105) at (2.25, 4.5) {$\otimes$};
		\node [style=none] (106) at (6.25, 0.5) {};
		\node [style=none] (107) at (6.25, -0.5) {};
		\node [style=action] (108) at (7.75, -4.5) {$\otimes$};
		\node [style=action] (109) at (7.75, 4.5) {$\otimes$};
		\node [style=action] (110) at (-2, 1.5) {$\otimes$};
		\node [style=action] (111) at (-2, -1.5) {$\otimes$};
		\node [style=none] (112) at (-1, 0.5) {};
		\node [style=none] (113) at (-1, -0.5) {};
		\node [style=action] (114) at (3.5, 1.5) {$\otimes$};
		\node [style=action] (115) at (3.5, -1.5) {$\otimes$};
		\node [style=none] (116) at (4.5, 0.5) {};
		\node [style=none] (117) at (4.5, -0.5) {};
	\end{pgfonlayer}
	\begin{pgfonlayer}{edgelayer}
		\draw [style=midpoint1, bend right] (2) to (3);
		\draw [style=midpoint1] (3) to (1);
		\draw [style=midpoint1] (0) to (6);
		\draw [style=midpoint1] (6) to (2);
		\draw [style=midpoint1, bend left] (10) to (11);
		\draw [style=midpoint1] (11) to (9);
		\draw [style=midpoint1] (8) to (14);
		\draw [style=midpoint1] (14) to (10);
		\draw [style=midpoint1, bend right=45] (18.center) to (14);
		\draw [style=midpoint1, bend left=45] (16.center) to (6);
		\draw [style=midpoint1, bend right] (36) to (37);
		\draw [style=midpoint1, bend left] (43) to (44);
		\draw [style=midpoint1, bend right] (68) to (69);
		\draw [style=midpoint1, bend left] (75) to (76);
		\draw [style=midpointy1pointy, in=0, out=180, looseness=0.75] (100) to (99.center);
		\draw [style=midpointy1pointy, in=0, out=180, looseness=0.75] (101) to (98.center);
		\draw [style=midpointy1pointy, in=0, out=180, looseness=0.75] (104) to (103.center);
		\draw [style=midpointy1pointy, in=0, out=180, looseness=0.75] (105) to (102.center);
		\draw [in=0, out=-180] (99.center) to (16.center);
		\draw [in=0, out=-180] (98.center) to (18.center);
		\draw [style=midpointy1pointy, in=0, out=180, looseness=0.75] (108) to (107.center);
		\draw [style=midpointy1pointy, in=0, out=180, looseness=0.75] (109) to (106.center);
		\draw [style=midpoint1] (71) to (109);
		\draw [style=midpoint1] (109) to (69);
		\draw [style=midpoint1] (69) to (105);
		\draw [style=midpoint1] (105) to (37);
		\draw [style=midpoint1] (37) to (101);
		\draw [style=midpoint1] (101) to (3);
		\draw [style=midpoint1] (78) to (108);
		\draw [style=midpoint1] (108) to (76);
		\draw [style=midpoint1] (76) to (104);
		\draw [style=midpoint1] (104) to (44);
		\draw [style=midpoint1] (44) to (100);
		\draw [style=midpoint1] (100) to (11);
		\draw [style=midpoint1, bend right=45] (113.center) to (111);
		\draw [style=midpoint1, bend left=45] (112.center) to (110);
		\draw [in=0, out=-180] (103.center) to (112.center);
		\draw [in=0, out=-180] (102.center) to (113.center);
		\draw [style=midpoint1, bend right=45] (117.center) to (115);
		\draw [style=midpoint1, bend left=45] (116.center) to (114);
		\draw [in=0, out=-180] (107.center) to (116.center);
		\draw [in=0, out=-180] (106.center) to (117.center);
		\draw [style=midpoint1] (2) to (110);
		\draw [style=midpoint1] (110) to (36);
		\draw [style=midpoint1] (36) to (114);
		\draw [style=midpoint1] (114) to (68);
		\draw [style=midpoint1] (68) to (70);
		\draw [style=midpoint1] (10) to (111);
		\draw [style=midpoint1] (111) to (43);
		\draw [style=midpoint1] (43) to (115);
		\draw [style=midpoint1] (115) to (75);
		\draw [style=midpoint1] (75) to (77);
	\end{pgfonlayer}
\end{tikzpicture}
\]
Given a context with a strong prior about expected sensory states and a continuation that responds to an action of type \(A_0\) by feeding back a state on \(S_0\), the best response can be shown to be that which selects actions that, under the current state, maximize the likelihood of obtaining the expected `goal' state \citep{Buckley2017free,Friston2015Active}.
\end{ex}

\begin{rmk}
We have framed each of these statistical procedures as optimization problems not only to suggest a link to the utility-maximising agents of game theory, but also because it suggests the use of iterative methods to compute best responses; note that computational tractability is an important motivation in the proof of Proposition \ref{prop:vae-learns-model}. 

The question of providing such dynamical or, thinking of game composition as an algebra for building complex systems, `coalgebraic' semantics for (generalized) optimization games is the topic of the next section. We first formalize this notion.
\end{rmk}

\begin{defn} \label{def:opt-game}
An \textbf{optimization game} is any open game whose best response function can be defined by a function of the form \(\Sigma \times C \xto{\pi} M \xto{\varphi} P\), where \(\Sigma\) is a strategy type, \(C\) a context type, \(M\) any space, and \(P\) a poset. We call \(\varphi\) the \textbf{fitness function}, and think of \(\pi\) as projecting systems into a space whose points can be assigned a fitness. The best response function of an optimization game can then be defined by giving the subset of strategies contextually maximizing fitness, for each context \(c : C\).
\end{defn}

\section{Cybernetic Systems and Dynamical Realisation}
\label{sec:org4ff5161}
\label{sec:cyber-sys}

In this section, we begin to answer the question of precisely how the optimization games of the previous section may be realized in physical systems, such as brains or computers. More formally, this means we seek open dynamical systems whose input and output types correspond to the domain and codomain types of the foregoing games, such that there is a correspondence between the behaviours of the abstract games and their dynamical realisations, and such that the evolutions of the internal states of the dynamical systems correspond to strategic improvements in game-playing: by concentrating on optimization games, a natural measure of such improvement is encoded in the fitness function underlying the best-response relator.

We do not require that there is a correspondence between internal states of the realisations and strategies for the corresponding games, but we do require that the fitness functions extend to the the total state spaces of the closure of a realisation induced by the context. When there \emph{is} a correspondence between internal states and strategies, we can take advantage of Definition \ref{def:open-game} and interpret trajectories over the state space as trajectories over strategies witnessing the strategic improvement.

We begin by sketching categories of dynamical games, and then use these ideas to define preliminary notions of open cybernetic systems and categories thereof. We consider principally single systems whose underlying games are atomic (in the sense of Remark \ref{rmk:atomic-games}), and leave the study of the behaviour of interacting cybernetic systems to future work. Once more, we omit proofs in this section; they will appear in a paper to follow.

\begin{defn}[Discrete-time dynamical system over $\cat{C}$; after \citep{Schultz2019Dynamical,Clarke2020Profunctor}] \label{def:dds}
A \textbf{discrete-time dynamical system} over \(\cat{C}\) with state space \(S :\cat{C}\), input type \(A : \cat{C}\) and output type \(B : \cat{C}\) is a lens \((S, S) \lensto (B, A)\) over \(\cat{C}\), \emph{i.e.} in the following optical hom object:
\begin{align*}
\int^{M : \cat{C}} \Cat{Comon}(\cat{C})(S, M \otimes B) \times \cat{C}(M \otimes A,S)
\cong \Cat{Comon}(\cat{C})(S, B) \times \cat{C}(S \otimes A, S)
\end{align*}
where the isomorphism follows by Yoneda reduction. Note that this requires that the `output' map of the dynamical system is a comonoid homomorphism in \(\cat{C}\) and hence deterministic in a category of stochastic channels.
\end{defn}

\begin{defn}[Category of discrete-time dynamical systems] \label{def:cat-dds}
We define a category \(\Cat{Dyn}_{\cat{C}}\) whose objects are the objects of \(\cat{C}\) and whose morphisms, denoted \(A \xto{S} B\), are discrete-time dynamical systems; the symbol above the arrow denotes the internal state space. Hom objects are given by
\[
\Cat{Dyn}_{\cat{C}}(A, B) = \sum_{S : \cat{C}} \Cat{Comon}(\cat{C})(S, B) \times \cat{C}(S \otimes A, S) \, .
\]
Identity dynamical systems on each \(A : \cat{C}\) are the `no-op' dynamical systems \(A \xto{A} A\) given by identity optics \(\id_A : (A, A) \lensto (A, A)\). Associativity and unitality of composition is inherited from the category of optics underlying Definition \ref{def:dds}; a symmetric monoidal structure is similarly inherited. \qed
\end{defn}

\begin{defn}[Lenses over dynamical systems; after \citep{Riley2018Categories}] \label{def:dyn-lens}
The category of (monoidal) lenses over \(\cat{C}\text{-dynamical}\) systems has as objects pairs \((X, A)\) of objects in \(\cat{C}\) and as morphisms, \textbf{dynamical lenses} \((X, A) \lensto (Y, B)\), elements of the type
\begin{gather*}
\begin{tikzpicture}
	\begin{pgfonlayer}{nodelayer}
		\node [style=dot] (17) at (-3, 2) {$X$};
		\node [style=dot] (18) at (-3, -2) {$A$};
		\node [style=copier] (19) at (0, 2) {};
		\node [style=copier] (20) at (0, -2) {};
		\node [style=dot] (21) at (3, 2) {$Y$};
		\node [style=dot] (22) at (3, -2) {$B$};
	\end{pgfonlayer}
	\begin{pgfonlayer}{edgelayer}
		\draw [style=midpoint1] (17) to (19);
		\draw [style=midpoint1] (19) to (21);
		\draw [style=midpoint1] (22) to (20);
		\draw [style=midpoint1] (20) to (18);
		\draw [style=midpoint1, bend left] (19) to (20);
	\end{pgfonlayer}
\end{tikzpicture} \\
\int^{M : \cat{C}} \Cat{Dyn}_{\cat{C}}(X, M \otimes Y) \times \Cat{Dyn}_{\cat{C}}(M \otimes B, A) \\
\cong \\
\sum_{P,Q : \cat{C}} \int^{M : \cat{C}} \cat{C}(P \otimes X, P) \times \Cat{Comon}(\cat{C})(P, M \otimes Y) \times \cat{C}(Q \otimes M \otimes B, Q) \times \Cat{Comon}(\cat{C})(Q, A) \\
\sum_{P,Q : \cat{C}} \;\; 
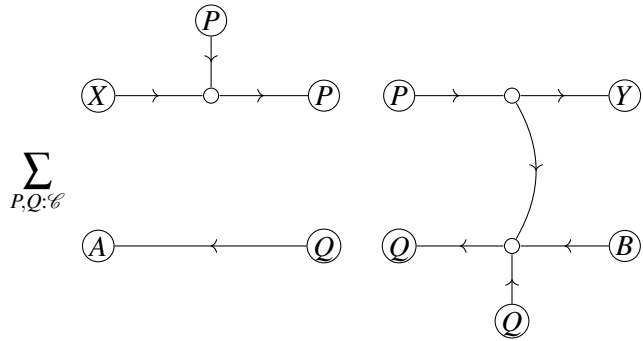
\begin{tikzpicture}
	\begin{pgfonlayer}{nodelayer}
		\node [style=dot] (17) at (-7, 2) {$X$};
		\node [style=dot] (18) at (-7, -2) {$A$};
		\node [style=copier] (19) at (-4, 2) {};
		\node [style=dot] (21) at (-1, 2) {$P$};
		\node [style=dot] (22) at (-1, -2) {$Q$};
		\node [style=dot] (24) at (1, 2) {$P$};
		\node [style=dot] (25) at (1, -2) {$Q$};
		\node [style=copier] (26) at (4, 2) {};
		\node [style=copier] (27) at (4, -2) {};
		\node [style=dot] (28) at (7, 2) {$Y$};
		\node [style=dot] (29) at (7, -2) {$B$};
		\node [style=dot] (30) at (-4, 4) {$P$};
		\node [style=dot] (31) at (4, -4) {$Q$};
	\end{pgfonlayer}
	\begin{pgfonlayer}{edgelayer}
		\draw [style=midpoint1] (17) to (19);
		\draw [style=midpoint1] (19) to (21);
		\draw [style=midpoint1] (24) to (26);
		\draw [style=midpoint1] (26) to (28);
		\draw [style=midpoint1] (29) to (27);
		\draw [style=midpoint1] (27) to (25);
		\draw [style=midpoint1] (22) to (18);
		\draw [style=midpoint1, bend left] (26) to (27);
		\draw [style=midpoint1] (30) to (19);
		\draw [style=midpoint1] (31) to (27);
	\end{pgfonlayer}
\end{tikzpicture} \; .
\end{gather*}
That is, a dynamical lens is a pair of dynamical systems coupled along some `residual' type.
\end{defn}

\begin{rmk}
At this point we begin to run into sizes issues. However, for the purposes of this paper, we will simply assume that a satisfactory resolution of these matters is at hand; for instance, that there is a hierarchy of Grothendieck universes such that the coends over (`large') sums in the preceding definition constitute accessible objects.
\end{rmk}

We now expand the definition of context in the dynamical setting. We will see that a dynamical context is simply a closure of an open dynamical system: that is, a `larger' system into which a `smaller' open dynamical system can plug such that the composite is a closed (but still uninitialized) system.

\begin{prop} \label{prop:dyn-ctx}
If \(I\) is terminal in \(\cat{C}\), a context for a dynamical lens \((X, A) \lensto (Y, B)\) is an element of the following type, denoted \(\tilde{C}\big((X, A), (Y, B)\big)\):
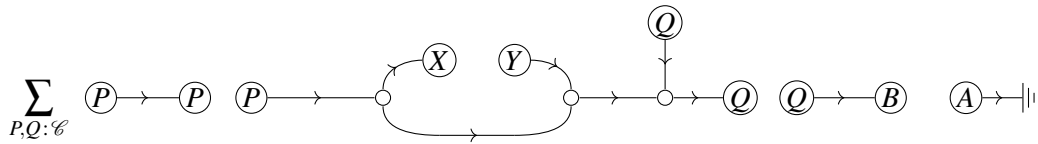
\[
\sum_{P,Q \, : \, \cat{C}} \;\; \begin{tikzpicture}
	\begin{pgfonlayer}{nodelayer}
		\node [style=copier] (0) at (-2.5, 0) {};
		\node [style=copier] (1) at (2.5, 0) {};
		\node [style=dot] (2) at (-1, 1) {$X$};
		\node [style=dot] (3) at (1, 1) {$Y$};
		\node [style=none] (4) at (-1, -1) {};
		\node [style=none] (5) at (1, -1) {};
		\node [style=dot] (6) at (-6, 0) {$P$};
		\node [style=dot] (7) at (7, 0) {$Q$};
		\node [style=dot] (8) at (13, 0) {$A$};
		\node [style=discarder0, rotate=0] (9) at (14.5, 0) {};
		\node [style=dot] (10) at (-10, 0) {$P$};
		\node [style=dot] (11) at (-7.5, 0) {$P$};
		\node [style=copier] (12) at (5, 0) {};
		\node [style=dot] (13) at (8.5, 0) {$Q$};
		\node [style=dot] (14) at (11, 0) {$B$};
		\node [style=dot] (15) at (5, 2) {$Q$};
	\end{pgfonlayer}
	\begin{pgfonlayer}{edgelayer}
		\draw [style=midpoint1] (6) to (0);
		\draw [style=midpoint1, in=-180, out=90] (0) to (2);
		\draw [style=midpoint1] (4.center) to (5.center);
		\draw [style=midpoint1, in=90, out=0] (3) to (1);
		\draw [in=-180, out=-90] (0) to (4.center);
		\draw [in=-90, out=0] (5.center) to (1);
		\draw [style=midpoint1] (8) to (9);
		\draw [style=midpoint1] (10) to (11);
		\draw [style=midpoint1] (1) to (12);
		\draw [style=midpoint1] (12) to (7);
		\draw [style=midpoint1] (13) to (14);
		\draw [style=midpoint1] (15) to (12);
	\end{pgfonlayer}
\end{tikzpicture}
\]
Interpreting this diagram, a context for a dynamical lens \((X, A) \lensto (Y, B)\) amounts to an autonomous dynamical system with output type of the form \(X \otimes M\) (for some residual type \(M\)), coupled along the residual \(M\) to an open dynamical system with input type \(Y \otimes M\) and output type \(B\); and the \(A\) type is discarded. This is precisely what we should expect from a dynamical analogue of Proposition \ref{prop:ctx-nice}.
\end{prop}

\begin{defn} \label{def:dyn-game}
A \textbf{dynamical game} is just a generalized open game (\ref{def:open-game}) over the category of dynamical lenses. We write \((X, A) \xto{\tilde{\Sigma}, S} (Y, B)\) to indicate both the strategy type \(\tilde{\Sigma}\) and state space \(S\). Dynamical games form a symmetric monoidal category in the corresponding way. For notational clarity, we will write \(\tilde{G}\) for a dynamical game, \(\tilde{P}\) for its play function, and \(\tilde{B}\) for its best response function.
\end{defn}

\begin{defn}[Dynamical realisation of an open game] \label{def:realisation}
Let \(G : (X, A) \xto{\Sigma} (Y, B)\) be an open game with \(X, A, Y, B\) all objects of some symmetric monoidal category \(\cat{C}\). A \textbf{dynamical realisation} of \(G\) is a choice of dynamical game \(\tilde{G} : (X, A) \xto{\tilde{\Sigma}, S} (Y, B)\) on the same objects, along with a function \(\interp{\cdot} : C((X, A), (Y, B)) \to \tilde{C}((X, A), (Y, B))\) lifting static contexts to dynamical contexts. Given a context \(\optar{\pi}{k} : C((X, A), (Y, B))\), we choose a representative \(\optar{\interp{\pi}}{\interp{k}} \cong \interp{\optar{\pi}{k}} : \tilde{C}((X, A), (Y, B))\) for its realisation.
\end{defn}

A `dynamical context' is an element of the type given in Proposition \(\ref{prop:dyn-ctx}\): a context for a dynamical lens. A `static context' is simply a context for the `static' game that is being dynamically realized. At this stage, we impose no particular requirements on the context realisation function \(\interp{\cdot}\), except to say that in the intended semantics, \(\interp{\optar{\pi}{k}}\) is a (coupled, open) dynamical system that constantly emits the state \(\pi\) and (by some mechanism) realizes the channel \(k\). We call such a context \emph{stationary} as neither \(\pi\) nor \(k\) vary in time; future work will generalize the results of this section to \emph{non-stationary} contexts.

\begin{defn}[Open cybernetic systems] \label{def:cyber-sys}
An open \textbf{cybernetic system} is defined by the data:
\begin{itemize}
\item an open optimization game (Def. \ref{def:opt-game}) \(G : (X, A) \xto{\Sigma} (Y, B)\) with \(X, A, Y, B\) all objects of some symmetric monoidal category \(\cat{C}\),
\item a fitness function \(\varphi_G : \Sigma \times C \to M \xto{\varphi} F\) where \(C = C\left((X, A), (Y, B)\right)\),
\item a dynamical realisation \(\big(\tilde{G} : (X, A) \xto{\tilde{\Sigma}, S} (Y, B), \interp{\cdot} : C((X, A), (Y, B)) \to \tilde{C}((X, A), (Y, B)) \big)\) of \(G\),
\end{itemize}
satisfying the following condition for each context \(\optar{\pi}{k} : C((X, A), (Y, B))\):
\begin{itemize}
\item there exists a dynamical strategy \(\tilde{\sigma} : \tilde{\Sigma}\), such that
\item writing \(Z\) for the total state space of the autonomous dynamical system \(\interp{\optar{\pi}{k}} \lenscirc \tilde{P}(\tilde{\sigma})\) induced by the context, there exists a function \(\nu : Z \to M\) projecting \(Z\) into the `fitness landscape' \(M\), such that
\item there exists a fitness-maximising fixed point \(\zeta^\ast : Z\), in the sense that
\item for some equilibrium strategy of the static system \(\sigma^\ast : \text{fix } B(\optar{\pi}{k})\), \(\varphi(\nu(\zeta^\ast)) \leq \varphi_G(\sigma^\ast, \optar{\pi}{k})\).
\end{itemize}
\noindent
A \textbf{category of open cybernetic systems} is a category of (generalized) open games such that each game is an open cybernetic system with dynamics realised in the same category \(\cat{C}\), and such that the composite of games is a cybernetic system whose fitness-maximising fixed point projects onto fitness-maximising fixed points of each of the factors in their corresponding local contexts. (See \citep[§3.7]{Bolt2019Bayesian} for the definition of local context.)
\end{defn}
The idea here is that, by using the fitness function of the underlying optimization game, the cybernetic condition forces the behaviour of the dynamical realisation to coincide with the process of iteratively improving the strategies deployed by the system in playing the game. We summarize the condition in the diagram
\begin{equation*}
\begin{tikzcd}
\Sigma \times C \arrow[d, "\interp{\cdot}"'] \arrow[r]  & M \arrow[r, "\varphi"] & F \\
\tilde{\Sigma} \times \tilde{C} \arrow[r, "\text{fix}"] & Z \arrow[u]            &  
\end{tikzcd}
\end{equation*}
though this is in general ill-defined: we do not require a function \(\interp{\cdot} : \Sigma \to \tilde{\Sigma}\), and nor do we require that the best response to \(\tilde{G}\) coincides in any way with the best reponse to \(G\). Investigating such conditions is the subject of future work; for instance, we may be interested in nested cybernetic systems, such as characterize evolution by natural selection, and how their fitness functions constrain one another. For similar reasons, we are also interested in the case where the fitness function is itself non-stationary.

\begin{rmk} \label{rmk:kubernetes}
The codomain category of the cybernetic realisation functor is in general much larger than the domain category of static games, and often it makes sense to consider dynamical games in this codomain category as if they were dynamical realisations of static games, even if in fact there is no static game to which they could correspond. For instance, adaptive systems in physical environments are in general not realisations of static games because their contexts are irreducibly dynamical and thus not the dynamical realisation of a static context; but over short time intervals, it can be productive to treat such systems as realisations of static games. In continuous time (not treated here), it is even possible to consider dynamical games that are indeed realisations of games that are static when represented in a smoothly varying coordinate system. The free-energy framework of Theorem \ref{thm:cyber-fep} is an example of a category of cybernetic systems with a rich underlying category of dynamic games.
\end{rmk}

A classic category of open cybernetic systems is found in the computational neuroscience literature, as summarized in the following theorem.

\begin{thm} \label{thm:cyber-fep}
Consider the subcategory of \(\Cat{BayesLens}\) spanned by finite-dimensional Euclidean spaces, with morphisms generated (under sequential and parallel composition) by the (variational) autoencoder and inference games whose forwards and backwards channels emit Gaussian measures with high-precision. The (discrete-time) free-energy framework for action and perception \citep{Buckley2017free} instantiates a category of open cybernetic systems realising games over this subcategory.
\end{thm}

\begin{rmk}
Typical presentations of `active inference' under the free-energy principle are excessively complicated by the lack of attention paid to compositionality. Because the free-energy framework instantiates a \emph{category} of open cybernetic systems, a radically simplified compositional presentation is possible. Such a presentation forms a companion to the present work.
\end{rmk}

\begin{cor} \label{cor:bx-cortex}
The free-energy framework has been used to supply a computational explanation for the pervasive bidirectionality of cortical circuits in the mammalian brain \citep{Bastos2012Canonical,Friston2010free}. A corollary of Theorem \ref{thm:cyber-fep} is that this bidirectionality is furthermore justified by the abstract structure of Bayesian inference and its dynamical realisation: because Bayesian updates compose optically, a cybernetic system realising Bayesian inference compositionally must instantiate this structure. We note also that the parallel interacting bidirectional structure of the active inference game (Example \ref{ex:ai-game}) is reproduced in the cortex.
\end{cor}

The free-energy framework realisation of autoencoder games is not unique; an alternative is found in machine learning.

\begin{thm} \label{thm:cyber-vae}
Consider the subcategory of \(\Cat{BayesLens}\) spanned by finite-dimensional Euclidean spaces, with morphisms generated (under sequential and parallel composition) by the (variational) autoencoder and inference games whose forwards and backwards channels emit exponential-family measures. The deep (variational) autoencoder framework \citep{Kingma2017Variational} instantiates a category of open cybernetic systems realising games over this subcategory.
\end{thm}

Increasingly, the variational autoencoder framework is used to model complete agents in machine learning, rather than merely dynamically realise static inference or learning problems. Indeed, thinking of the `free-energy framework' as a collection of cybernetic realisations of autoencoder and active-inference games, the demonstration of the following corollary of Theorem \ref{thm:cyber-vae} is unsurprising:

\begin{cor} \label{cor:deep-ai}
The ``deep active inference agent'' \citep{Ueltzhoeffer2018Deep} is a cybernetic system realising an active inference game in the variational autoencoder framework.
\end{cor}

We have heretofore concentrated on `variational Bayesian' realisations of the games introduced in \secref{sec:games}, as they most strikingly fit the language of optimization used there. But we expect any other family of approximate inference methods to supply a corresponding category of cybernetic systems. We thus make the following conjecture.

\begin{conjecture}
Consider the subcategory of \(\Cat{BayesLens}\) spanned by finite-dimensional smooth manifolds, with morphisms generated (under sequential and parallel composition) by the generalized autoencoder and inference games. We expect sampling algorithms, such as Markov chain Monte Carlo, to supply a corresponding category of open cybernetic systems of interest.
\end{conjecture}

Finally, we provide further justification for Remark \ref{rmk:b-relt}.

\begin{obs} \label{prop:sigma-traj}
Consider a variational autoencoder, realised as in Theorem \ref{thm:cyber-vae}. By choosing the parameterizations \(\mathcal{F},\mathcal{P}\) of the forwards and backwards channels to coincide with the state spaces of their dynamical realisations, and the (static) play function \(P\) to take a parameter vector to the corresponding channel, the dynamical realisation induces a trajectory over the strategy space. Such trajectories organize into sheaf whose sections are trajectories of arbitrary length \citep{Schultz2019Dynamical}, spans of which are again just (generalized) dynamical systems; these spans are equivalently profunctors \citep{Benabou2000Distributors}. We can thus define a best-response function valued in profunctors whose elements are trajectories witnessing deviations of strategies to `better' strategies, and whose dynamical equilibria correspond precisely to the equilibria of the `static' best response function.
\end{obs}

\paragraph{On-going and Future Work}
\label{sec:org420ea85}

The structures sketched in this paper are merely first steps towards a categorical theory of cybernetics. In particular, since the first draft of this work was written, we have come to believe that the preliminary notions presented here of dynamical realisation, and by extension of open cybernetic system, are substantially less elegant than they could be. On-going work is focusing on this issue. We hope that a consequence of this refinement will be that the treatment of \emph{interacting} cybernetic systems is simplified. In this new setting, we will also treat non-stationary systems in dynamical contexts and in continuous time, thereby supplying a general compositional treatment of (amongst other things) the `free-energy' framework.

Finally, with respect to applications, we are interested in using these tools to realise game-theoretic games and to investigate the connections between repeated games and dynamical realisation. There are deep links with reinforcement learning to be explored, and we seek a setting for the study of nested and mutli-agent (`ecological') systems.

\bibliographystyle{eptcs}
\bibliography{bibliography}

\begin{thebibliography}{10}
\providecommand{\bibitemdeclare}[2]{}
\providecommand{\surnamestart}{}
\providecommand{\surnameend}{}
\providecommand{\urlprefix}{Available at }
\providecommand{\url}[1]{\texttt{#1}}
\providecommand{\href}[2]{\texttt{#2}}
\providecommand{\urlalt}[2]{\href{#1}{#2}}
\providecommand{\doi}[1]{doi:\urlalt{http://dx.doi.org/#1}{#1}}
\providecommand{\bibinfo}[2]{#2}

\bibitemdeclare{article}{Bastos2012Canonical}
\bibitem{Bastos2012Canonical}
\bibinfo{author}{A.~M. \surnamestart Bastos\surnameend}, \bibinfo{author}{W.~M.
  \surnamestart Usrey\surnameend}, \bibinfo{author}{R.~A. \surnamestart
  Adams\surnameend}, \bibinfo{author}{G.~R. \surnamestart Mangun\surnameend},
  \bibinfo{author}{P.~\surnamestart Fries\surnameend} \& \bibinfo{author}{K.~J.
  \surnamestart Friston\surnameend} (\bibinfo{year}{2012}):
  \emph{\bibinfo{title}{Canonical microcircuits for predictive coding}}.
\newblock {\sl \bibinfo{journal}{Neuron}}
  \bibinfo{volume}{76}(\bibinfo{number}{4}), pp. \bibinfo{pages}{695--711},
  \doi{10.1016/j.neuron.2012.10.038}.

\bibitemdeclare{article}{Bolt2019Bayesian}
\bibitem{Bolt2019Bayesian}
\bibinfo{author}{Joe \surnamestart Bolt\surnameend}, \bibinfo{author}{Jules
  \surnamestart Hedges\surnameend} \& \bibinfo{author}{Philipp \surnamestart
  Zahn\surnameend} (\bibinfo{year}{2019}): \emph{\bibinfo{title}{Bayesian open
  games}}.
\newblock \urlprefix\url{http://arxiv.org/abs/1910.03656v1}.

\bibitemdeclare{article}{Buckley2017free}
\bibitem{Buckley2017free}
\bibinfo{author}{Christopher~L \surnamestart Buckley\surnameend},
  \bibinfo{author}{Chang~Sub \surnamestart Kim\surnameend},
  \bibinfo{author}{Simon \surnamestart McGregor\surnameend} \&
  \bibinfo{author}{Anil~K \surnamestart Seth\surnameend}
  (\bibinfo{year}{2017}): \emph{\bibinfo{title}{The free energy principle for
  action and perception: A mathematical review}}.
\newblock {\sl \bibinfo{journal}{Journal of Mathematical Psychology}}
  \bibinfo{volume}{81}, pp. \bibinfo{pages}{55--79},
  \doi{10.1016/j.jmp.2017.09.004}.
\newblock \urlprefix\url{http://arxiv.org/abs/1705.09156v1}.

\bibitemdeclare{unpublished}{Benabou2000Distributors}
\bibitem{Benabou2000Distributors}
\bibinfo{author}{Jean \surnamestart Bénabou\surnameend}
  (\bibinfo{year}{2000}): \emph{\bibinfo{title}{Distributors at work}}.
\newblock
  \urlprefix\url{http://www.mathematik.tu-darmstadt.de/~streicher/FIBR/DiWo.pdf}.
\newblock \bibinfo{note}{Lecture notes written by Thomas Streicher}.

\bibitemdeclare{article}{Cho2017Disintegration}
\bibitem{Cho2017Disintegration}
\bibinfo{author}{Kenta \surnamestart Cho\surnameend} \& \bibinfo{author}{Bart
  \surnamestart Jacobs\surnameend} (\bibinfo{year}{2017}):
  \emph{\bibinfo{title}{Disintegration and Bayesian Inversion via String
  Diagrams}}.
\newblock {\sl \bibinfo{journal}{Math. Struct. Comp. Sci. 29 (2019) 938-971}},
  \doi{10.1017/S0960129518000488}.
\newblock \urlprefix\url{http://arxiv.org/abs/1709.00322v3}.

\bibitemdeclare{article}{Clarke2020Profunctor}
\bibitem{Clarke2020Profunctor}
\bibinfo{author}{Bryce \surnamestart Clarke\surnameend}, \bibinfo{author}{Derek
  \surnamestart Elkins\surnameend}, \bibinfo{author}{Jeremy \surnamestart
  Gibbons\surnameend}, \bibinfo{author}{Fosco \surnamestart
  Loregian\surnameend}, \bibinfo{author}{Bartosz \surnamestart
  Milewski\surnameend}, \bibinfo{author}{Emily \surnamestart
  Pillmore\surnameend} \& \bibinfo{author}{Mario \surnamestart
  Román\surnameend} (\bibinfo{year}{2020}): \emph{\bibinfo{title}{Profunctor
  optics, a categorical update}}.
\newblock \urlprefix\url{http://arxiv.org/abs/2001.07488v1}.

\bibitemdeclare{article}{Coecke2016Categorical}
\bibitem{Coecke2016Categorical}
\bibinfo{author}{Bob \surnamestart Coecke\surnameend} \& \bibinfo{author}{Aleks
  \surnamestart Kissinger\surnameend} (\bibinfo{year}{2016}):
  \emph{\bibinfo{title}{Categorical Quantum Mechanics II: Classical-Quantum
  Interaction}}.
\newblock \doi{10.1142/S0219749910006502}.
\newblock \urlprefix\url{http://arxiv.org/abs/1605.08617v1}.

\bibitemdeclare{incollection}{Coecke2015Categorical}
\bibitem{Coecke2015Categorical}
\bibinfo{author}{Bob \surnamestart Coecke\surnameend} \& \bibinfo{author}{Aleks
  \surnamestart Kissinger\surnameend} (\bibinfo{year}{2017}):
  \emph{\bibinfo{title}{Categorical Quantum Mechanics I: Causal Quantum
  Processes}}.
\newblock In \bibinfo{editor}{Elaine \surnamestart Landry\surnameend}, editor:
  {\sl \bibinfo{booktitle}{Categories for the Working Philosopher}},
  chapter~\bibinfo{chapter}{12}, \bibinfo{publisher}{Oxford University Press},
  pp. \bibinfo{pages}{286--328}.
\newblock \urlprefix\url{https://arxiv.org/abs/1510.05468v3}.

\bibitemdeclare{article}{Foster2007Combinators}
\bibitem{Foster2007Combinators}
\bibinfo{author}{J.~Nathan \surnamestart Foster\surnameend},
  \bibinfo{author}{Michael~B. \surnamestart Greenwald\surnameend},
  \bibinfo{author}{Jonathan~T. \surnamestart Moore\surnameend},
  \bibinfo{author}{Benjamin~C. \surnamestart Pierce\surnameend} \&
  \bibinfo{author}{Alan \surnamestart Schmitt\surnameend}
  (\bibinfo{year}{2007}): \emph{\bibinfo{title}{Combinators for bidirectional
  tree transformations}}.
\newblock {\sl \bibinfo{journal}{{ACM} Transactions on Programming Languages
  and Systems}} \bibinfo{volume}{29}(\bibinfo{number}{3}),
  p.~\bibinfo{pages}{17}, \doi{10.1145/1232420.1232424}.

\bibitemdeclare{article}{Friston2010free}
\bibitem{Friston2010free}
\bibinfo{author}{Karl \surnamestart Friston\surnameend} (\bibinfo{year}{2010}):
  \emph{\bibinfo{title}{The free-energy principle: a unified brain theory?}}
\newblock {\sl \bibinfo{journal}{Nature Reviews Neuroscience}}
  \bibinfo{volume}{11}(\bibinfo{number}{2}), pp. \bibinfo{pages}{127--138},
  \doi{10.1038/nrn2787}.

\bibitemdeclare{article}{Friston2015Active}
\bibitem{Friston2015Active}
\bibinfo{author}{Karl \surnamestart Friston\surnameend},
  \bibinfo{author}{Francesco \surnamestart Rigoli\surnameend},
  \bibinfo{author}{Dimitri \surnamestart Ognibene\surnameend},
  \bibinfo{author}{Christoph \surnamestart Mathys\surnameend},
  \bibinfo{author}{Thomas \surnamestart Fitzgerald\surnameend} \&
  \bibinfo{author}{Giovanni \surnamestart Pezzulo\surnameend}
  (\bibinfo{year}{2015}): \emph{\bibinfo{title}{Active inference and epistemic
  value}}.
\newblock {\sl \bibinfo{journal}{Cognitive Neuroscience}}
  \bibinfo{volume}{6}(\bibinfo{number}{4}), pp. \bibinfo{pages}{187--214},
  \doi{10.1080/17588928.2015.1020053}.

\bibitemdeclare{article}{Fritz2019synthetic}
\bibitem{Fritz2019synthetic}
\bibinfo{author}{Tobias \surnamestart Fritz\surnameend} (\bibinfo{year}{2019}):
  \emph{\bibinfo{title}{A synthetic approach to Markov kernels, conditional
  independence and theorems on sufficient statistics}}.
\newblock \urlprefix\url{http://arxiv.org/abs/1908.07021v8}.

\bibitemdeclare{article}{Ghani2016Compositional}
\bibitem{Ghani2016Compositional}
\bibinfo{author}{Neil \surnamestart Ghani\surnameend}, \bibinfo{author}{Jules
  \surnamestart Hedges\surnameend}, \bibinfo{author}{Viktor \surnamestart
  Winschel\surnameend} \& \bibinfo{author}{Philipp \surnamestart
  Zahn\surnameend} (\bibinfo{year}{2016}): \emph{\bibinfo{title}{Compositional
  game theory}}.
\newblock {\sl \bibinfo{journal}{Proceedings of Logic in Computer Science
  (LiCS) 2018}}, \doi{10.1145/3209108.3209165}.
\newblock \urlprefix\url{http://arxiv.org/abs/1603.04641v3}.

\bibitemdeclare{article}{Heunen2017Convenient}
\bibitem{Heunen2017Convenient}
\bibinfo{author}{Chris \surnamestart Heunen\surnameend}, \bibinfo{author}{Ohad
  \surnamestart Kammar\surnameend}, \bibinfo{author}{Sam \surnamestart
  Staton\surnameend} \& \bibinfo{author}{Hongseok \surnamestart
  Yang\surnameend} (\bibinfo{year}{2017}): \emph{\bibinfo{title}{A Convenient
  Category for Higher-Order Probability Theory}}.
\newblock \doi{10.1109/lics.2017.8005137}.

\bibitemdeclare{}{Kingma2017Variational}
\bibitem{Kingma2017Variational}
\bibinfo{author}{Diederik~P. \surnamestart Kingma\surnameend}
  (\bibinfo{year}{2017}): \emph{\bibinfo{title}{Variational Inference \& Deep
  Learning}}.
\newblock
  \urlprefix\url{https://hdl.handle.net/11245.1/8e55e07f-e4be-458f-a929-2f9bc2d169e8}.

\bibitemdeclare{article}{Knoblauch2019Generalized}
\bibitem{Knoblauch2019Generalized}
\bibinfo{author}{Jeremias \surnamestart Knoblauch\surnameend},
  \bibinfo{author}{Jack \surnamestart Jewson\surnameend} \&
  \bibinfo{author}{Theodoros \surnamestart Damoulas\surnameend}
  (\bibinfo{year}{2019}): \emph{\bibinfo{title}{Generalized Variational
  Inference}}.
\newblock \urlprefix\url{http://arxiv.org/abs/1904.02063v4}.

\bibitemdeclare{article}{Loregian2015This}
\bibitem{Loregian2015This}
\bibinfo{author}{Fosco \surnamestart Loregian\surnameend}
  (\bibinfo{year}{2015}): \emph{\bibinfo{title}{This is the (co)end, my only
  (co)friend}}.
\newblock \urlprefix\url{http://arxiv.org/abs/1501.02503v4}.

\bibitemdeclare{incollection}{Mellies2006Functorial}
\bibitem{Mellies2006Functorial}
\bibinfo{author}{Paul-Andr{\'{e}} \surnamestart Melli{\`{e}}s\surnameend}
  (\bibinfo{year}{2006}): \emph{\bibinfo{title}{Functorial Boxes in String
  Diagrams}}.
\newblock In: {\sl \bibinfo{booktitle}{Computer Science Logic}},
  \bibinfo{publisher}{Springer Berlin Heidelberg}, pp. \bibinfo{pages}{1--30},
  \doi{10.1007/11874683_1}.

\bibitemdeclare{article}{Moeller2018Monoidal}
\bibitem{Moeller2018Monoidal}
\bibinfo{author}{Joe \surnamestart Moeller\surnameend} \&
  \bibinfo{author}{Christina \surnamestart Vasilakopoulou\surnameend}
  (\bibinfo{year}{2018}): \emph{\bibinfo{title}{Monoidal Grothendieck
  Construction}}.
\newblock \urlprefix\url{http://arxiv.org/abs/1809.00727v2}.

\bibitemdeclare{misc}{nLab2020Grothendieck}
\bibitem{nLab2020Grothendieck}
\bibinfo{author}{\surnamestart {nLab authors}\surnameend}
  (\bibinfo{year}{2020}): \emph{\bibinfo{title}{Grothendieck construction}}.
\newblock
  \urlprefix\url{http://ncatlab.org/nlab/show/Grothendieck+construction}.
\newblock \bibinfo{note}{Revision 62}.

\bibitemdeclare{article}{Riley2018Categories}
\bibitem{Riley2018Categories}
\bibinfo{author}{Mitchell \surnamestart Riley\surnameend}
  (\bibinfo{year}{2018}): \emph{\bibinfo{title}{Categories of Optics}}.
\newblock \urlprefix\url{http://arxiv.org/abs/1809.00738v2}.

\bibitemdeclare{article}{Roman2020Open}
\bibitem{Roman2020Open}
\bibinfo{author}{Mario \surnamestart Román\surnameend} (\bibinfo{year}{2020}):
  \emph{\bibinfo{title}{Open Diagrams via Coend Calculus}}.
\newblock \urlprefix\url{http://arxiv.org/abs/2004.04526v2}.

\bibitemdeclare{article}{Roman2020Profunctor}
\bibitem{Roman2020Profunctor}
\bibinfo{author}{Mario \surnamestart Román\surnameend} (\bibinfo{year}{2020}):
  \emph{\bibinfo{title}{Profunctor optics and traversals}}.
\newblock \urlprefix\url{http://arxiv.org/abs/2001.08045v1}.

\bibitemdeclare{article}{Schultz2019Dynamical}
\bibitem{Schultz2019Dynamical}
\bibinfo{author}{Patrick \surnamestart Schultz\surnameend},
  \bibinfo{author}{David~I \surnamestart Spivak\surnameend} \&
  \bibinfo{author}{Christina \surnamestart Vasilakopoulou\surnameend}
  (\bibinfo{year}{2019}): \emph{\bibinfo{title}{Dynamical Systems and
  Sheaves}}.
\newblock {\sl \bibinfo{journal}{Applied Categorical Structures}}, pp.
  \bibinfo{pages}{1--57}, \doi{10.1007/s10485-019-09565-x}.
\newblock \urlprefix\url{http://arxiv.org/abs/1609.08086v4}.

\bibitemdeclare{article}{Smithe2020Bayesian}
\bibitem{Smithe2020Bayesian}
\bibinfo{author}{Toby St.~Clere \surnamestart Smithe\surnameend}
  (\bibinfo{year}{2020}): \emph{\bibinfo{title}{Bayesian Updates Compose
  Optically}}.
\newblock \urlprefix\url{http://arxiv.org/pdf/2006.01631v1}.

\bibitemdeclare{article}{Spivak2019Generalized}
\bibitem{Spivak2019Generalized}
\bibinfo{author}{David~I. \surnamestart Spivak\surnameend}
  (\bibinfo{year}{2019}): \emph{\bibinfo{title}{Generalized Lens Categories via
  functors $\mathcal{C}^\text{op} \to \mathbf{Cat}$}}.
\newblock \urlprefix\url{http://arxiv.org/abs/1908.02202v2}.

\bibitemdeclare{article}{Ueltzhoeffer2018Deep}
\bibitem{Ueltzhoeffer2018Deep}
\bibinfo{author}{Kai \surnamestart Ueltzhöffer\surnameend}
  (\bibinfo{year}{2018}): \emph{\bibinfo{title}{Deep Active Inference}}.
\newblock {\sl \bibinfo{journal}{Biological Cybernetics}}
  \bibinfo{volume}{112}(\bibinfo{number}{6}), pp. \bibinfo{pages}{547--573},
  \doi{10.1007/s00422-018-0785-7}.
\newblock \urlprefix\url{http://arxiv.org/abs/1709.02341v5}.

\end{thebibliography}

\end{document}